\documentclass[sigconf, nonacm]{acmart}

\usepackage{graphicx}
\usepackage{balance}
\usepackage{xcolor}
\usepackage{subfig}
\usepackage{verbatim}
\usepackage{caption}
\usepackage{multirow}
\usepackage{algorithm}
\usepackage{algpseudocode}
\usepackage{comment}
\usepackage{makecell}
\usepackage{colorprofiles}
\usepackage[a-2b, mathxmp]{pdfx}[2018/12/22]
\usepackage{hyperref}

\usepackage{float}
\floatstyle{plaintop}
\restylefloat{table}

\begin{document}
\title{Motiflets - Simple and Accurate Detection of Motifs in Time Series}

\author{Patrick Sch\"afer}
\affiliation{%
  \institution{Humboldt-Universit\"at zu Berlin}
  \city{Berlin}
  \country{Germany}
}
\email{patrick.schaefer@hu-berlin.de}

\author{Ulf Leser}
\affiliation{%
  \institution{Humboldt-Universit\"at zu Berlin}
  \city{Berlin}
  \country{Germany}
}
\email{leser@informatik.hu-berlin.de}

\sloppy
\begin{abstract}
A time series motif intuitively is a short time series that repeats itself approximately the same within a larger time series. Such motifs often represent concealed structures, such as heart beats in an ECG recording, the riff in a pop song, or sleep spindles in EEG sleep data. Motif discovery (MD) is the task of finding such motifs in a given input series. As there are varying definitions of what exactly a motif is, a number of different algorithms exist. As central parameters they all take the length $l$ of the motif and the maximal distance $r$ between the motif's occurrences. In practice, however, especially suitable values for $r$ are very hard to determine upfront, and  found motifs show a high variability even for very similar $r$ values. Accordingly, finding an interesting motif with these methods requires extensive trial-and-error.

In this paper, we present a different approach to the MD problem. We define $k$-Motiflets as the set of exactly $k$ occurrences of a motif of length $l$, whose maximum pairwise distance is minimal. This turns the MD problem upside-down: The central parameter of our approach is not the distance threshold $r$, but the desired number of occurrence $k$ of the motif, which we show is considerably more intuitive and easier to set. Based on this definition, we present exact and approximate algorithms for finding $k$-Motiflets and analyze their complexity. To further ease the use of our method, we describe statistical tools to automatically determine meaningful values for its input parameters. Thus, for the first time, extracting meaningful motif sets without any a-priori knowledge becomes feasible.
By evaluation on several real-world data sets and comparison to four state-of-the-art MD algorithms, we show that our proposed algorithm is both quantitatively superior to its competitors, finding larger motif sets at higher similarity, and qualitatively better, leading to clearer and easier to interpret motifs without any need for manual tuning.
\end{abstract}

\maketitle

\section{Introduction}
Time series (TS) are sequences of real values ordered along a specific dimension, with time as the most important dimension. The concept of time series motif discovery (TSMD, or MD in short) was first described in~\cite{patel2002mining} and has since then emerged as an important primitive for exploring and analyzing TS in the data mining and management community~\cite{linardi2018matrix, patel2002mining, mueen2009exact, grabocka2016latent, torkamani2017survey,yeh2016matrix}. Intuitively, MD is the problem of finding patterns, i.e., short TS, that repeat themselves approximately the same within a given TS. These motifs often reflect concealed structures in the process generating the TS, such as heart beats in an ECG recording~\cite{petrutiu2007abrupt} or sleep spindles and k-Complexes in EEG sleep data~\cite{kohlmorgen2000identification}. Applications of MD exist across many domains, such as seismic signals~\cite{siddiquee2019seismo}, electric household devices~\cite{shao2012temporal}, DNA sequences~\cite{eden2007discovering}, electrocardiography data~\cite{liu2015efficient}, wind generation turbines~\cite{kamath2012finding}, or audio signal analysis~\cite{gomes2015classifying}. MD is also important as a pre-processing step for classification, clustering, anomaly detection, and rule discovery in TS~\cite{torkamani2017survey}. For example, identified motifs can help to speed up feature extraction in TS classification~\cite{lee2018deepfinder}.

Though intuitively easy to describe, the specific definitions of the MD problem for a TS $T$ differ notably between existing works. Several tools focus only on \emph{motif pairs}~\cite{mueen2009exact,yeh2016matrix}, which are defined as the most similar pair(s) of subsequences of $T$ of user-defined length $l$. However, real-world motifs typically do not only occur in pairs; for example, heartbeats in ECG recordings are all similar to each other. A more general and arguably more natural approach to MD is the search for \emph{motif sets}, defined as the largest set of short TS approximately contained in $T$ and in \emph{some sense} close to each other. At least four different definitions exist for this \emph{in some sense}, namely (by date of publication): k-Motifs~\cite{patel2002mining}, Range Motif (RM)~\cite{mueen2009exact}, Learning Motif (LM)~\cite{grabocka2016latent}, and VALMOD Motif Sets~\cite{linardi2018valmod} (for precise definitions, see Section~\ref{sec:definitions}). All of these methods require users to provide two central parameters: the motif length $l$, and a distance threshold $r$. While the former can often be estimated using domain knowledge, the latter is very hard to set. Yet, no algorithms are known for learning the input parameters from the data.

In this paper, we introduce $k$-Motiflets, a novel definition for MD that turns the problem upside-down. $k$-Motiflets take the desired motif set size $k$ as parameter and maximize the similarity of the motif set. As we will show, this $k$ is an integer with an easily understood interpretation, and in many use cases the expected size of the motif set is known prior to the analysis. Consider for example the possible copyright fraud in the pop song “Ice Ice Baby” by Vanilla Ice compared to “Under pressure” by Queen / David Bowie. Listening to these songs it is easy to get a first estimate of the number of repetitions (parameter $k$) of the problematic sections. On the other hand, it is impossible for humans to guess a good value for real-valued distance between different repetitions (parameter $r$).

We argue that guessing k is almost always easier, as the concept of \emph{how many repetitions of a motif do you expect} is much easier to understand - though the guess itself need not be easy, and thus we will also offer algorithms to learn $k$. Furthermore, as $k$ is an integer, there is only a very limited number of options, as use cases with thousands of motif occurrences are rare. In contrast, the concept of \emph{how far apart do you expect motifs to be at maximum} is extremely difficult to understand as distances, e.g. Euclidean distance, are measured by an opaque mathematical formula for which no intuition exists. Furthermore, $r$ is a real value with infinitely many values, and even small changes may lead to gross changes in motifs found.

\begin{figure}[t]
	\includegraphics[width=1.0\columnwidth]{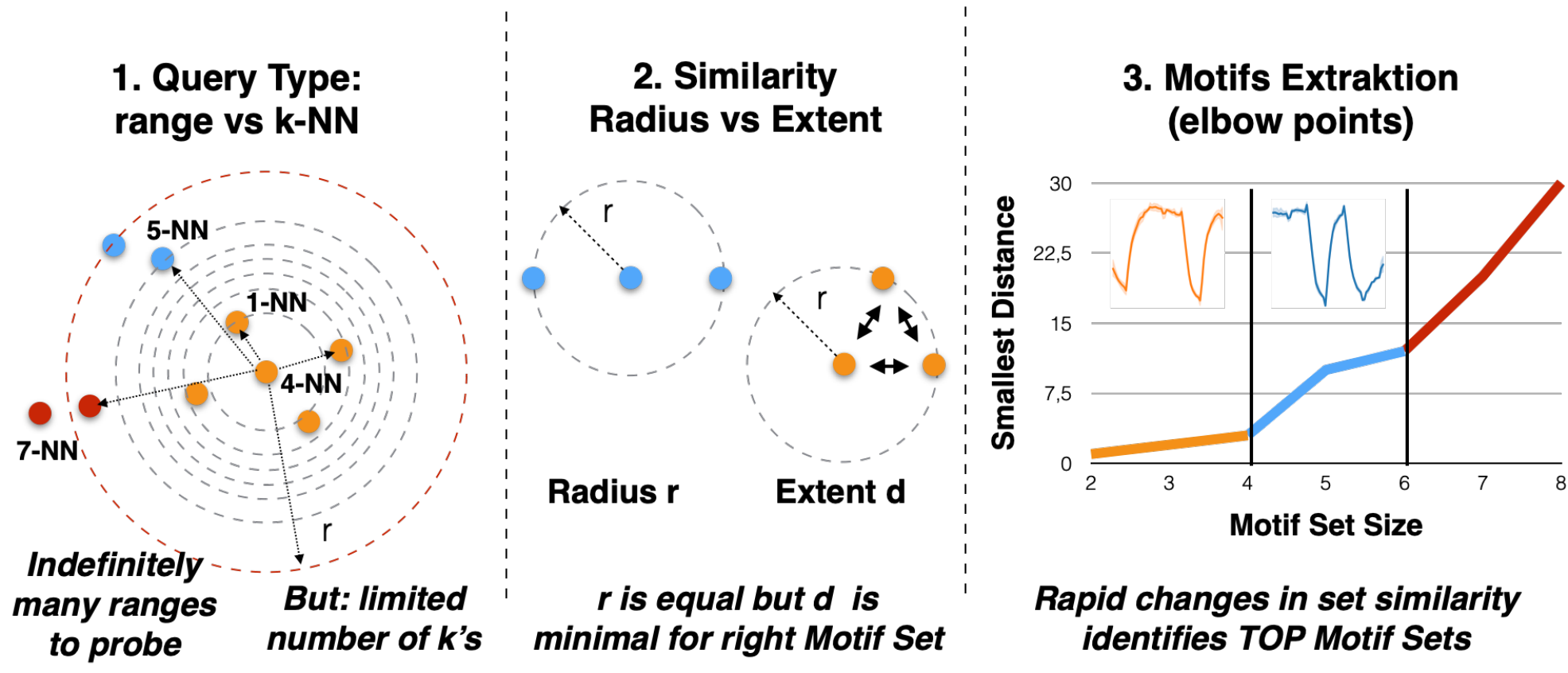}
 	\caption{Three major changes in k-Motiflets vs SotA: (1) the use of k-NN vs range queries, (2) minimizing pairwise distances vs the radius, and (3) guided extraction of meaningful motifs using the elbow plot.
	\label{fig:motiflets-vs-sota}
	}
\end{figure}
\begin{figure}[t]
    \centering
     	\includegraphics[width=0.9\columnwidth]{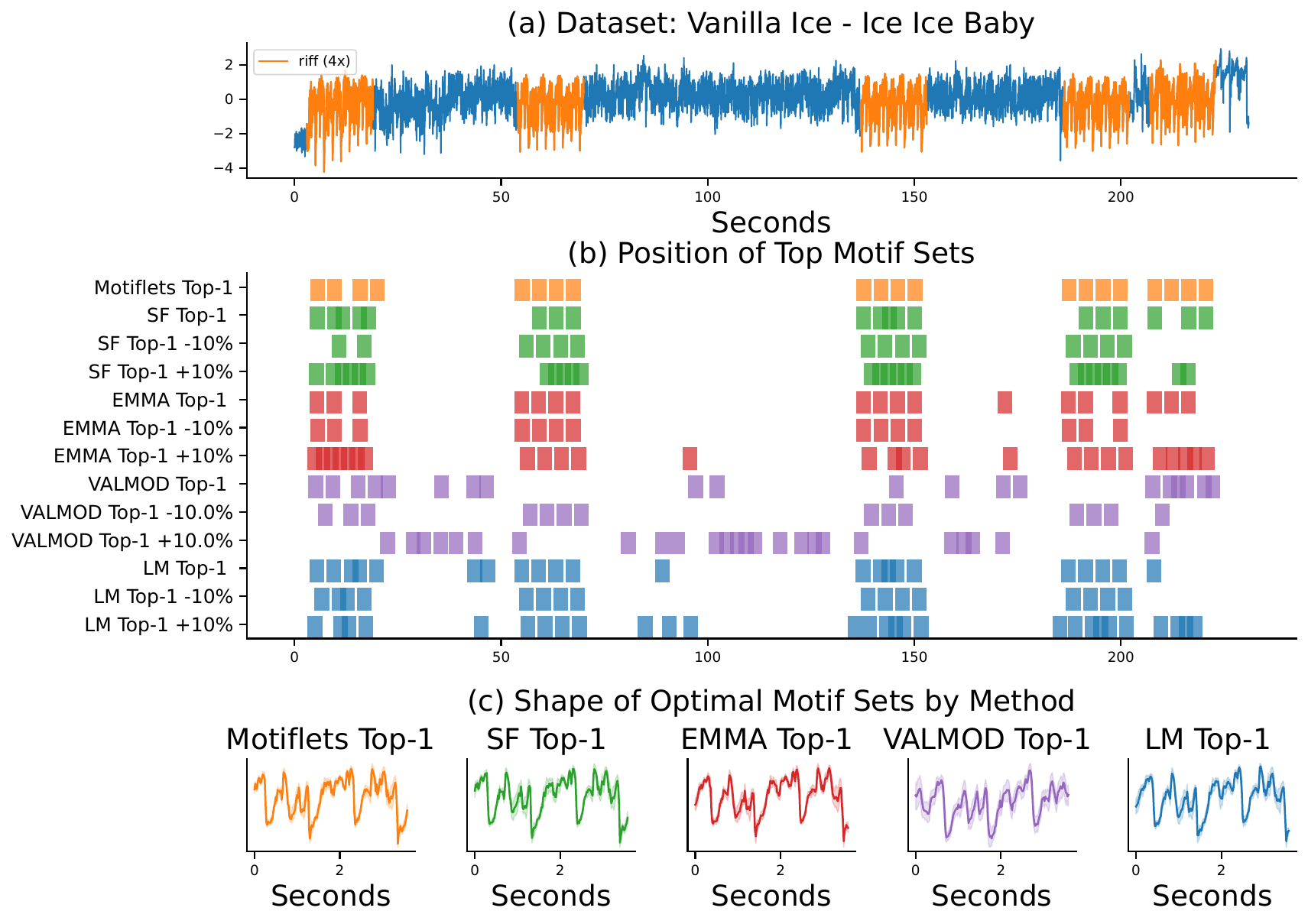}
        \caption{
        A comparison of MD definitions on the pop song Ice Ice Baby by Vanilla Ice. The TS contains one significant motif: the famous riff has $20$ repeats in $5$ blocks of $4$.
        From the competitors, none correctly identifies all occurrences, even when given optimal parameters. When adding a small noise, blurred or too small motifs are reported. Only our approximate $k$-Motiflets identify all $k=20$ repeats of the riff.}
    	\label{fig:pop-song}
\end{figure}

We illustrate the three main contributions in Figure~\ref{fig:motiflets-vs-sota}.
\emph{Firstly (Left)}: SotA is based on range queries, with radius $r$ as input. There are indefinitely many ranges to probe, whereas most return the same motif. A simple, yet overseen improvement to only get distinct motifs, is to use k-NN queries.
\emph{Secondly (Center)}: Consider two motif sets with 3 subsequences each. While both have the same radius, the right one has higher similarity. We use the maximal pairwise distance, called \emph{extent}, to find the TOP motif among all k-NN queries.
\emph{Finally (Right)}: We introduce elbow plots for a guided extraction of meaningful motif set sizes. Here, rapid changes in similarity when increasing k represent a characteristic change from one motif to another. Overall, we will show that these improvements reduce the runtime and human efforts to find motif sets considerably.
Consider for example Figure~\ref{fig:pop-song}. (Top) shows a TS extracted from the pop song Ice Ice Baby by Vanilla Ice using the 2nd MFCC channel sampled at 100Hz~\cite{yeh2016matrix}. This TS is a particularly famous pop song, as it is alleged to have copied its riff from "Under Pressure" by Queen and David Bowie. It contains $20$ repeats of the riff in $5$ blocks with each riff being $3.6-4$s long.  We applied our novel $k$-Motiflets to this problem and compared results to those of VALMOD~\cite{linardi2018valmod}, two implementations of k-Motifs - namely EMMA~\cite{lonardi2002finding} and Set Finder (SF)~\cite{bagnall2014finding} - and Learning Motifs (LM)~\cite{grabocka2016latent}).
We first ran $k$-Motiflets with $k=20$ and $l=3.6$s. From the found motif set, we inferred the radius $r$. We then ran all competitors with this exact $r$ value but also with some noise $\epsilon$ added on top ($ \epsilon \in \{-10\%, +10\%\}$) to reflect a typical trial-and-error scenario based on visual inspection of the data. Note that in this setup competitors are provided with inputs with near-optimal parameters, which are otherwise hard to guess.
$k$-Motiflets (orange squares in Figure~\ref{fig:set_motif_example} (b)) identified all $20$ riffs. None of the competitors found all occurrences though returning motif sets of size $20$. Instead, competitors found subsets of the riff, but their accuracy depends heavily on the precise parameterization. Already a slight deviation in these parameters (which must be manually set) leads to completely different motifs for all competitors; with increasing deviation, the motif suddenly covers the entire TS, and with decreasing values, the number of found occurrences shrinks considerably.
In summary, the contributions of this paper are as follows:
\begin{enumerate}
    \item We define $k$-Motiflets, a novel definition for MD in TS. In contrast to all prior works, this definition is based on the desired size $k$ of the motif set, not the maximum distance $r$ between occurrences of a motif.
    \item We present exact and approximate algorithms for finding $k$-Motiflets. As $k$-Motiflet can be considered as an extension of the Range Motif definition of motif sets, we thereby also provide the first implementation of this definition.
    \item We analyze the complexity of both algorithms and show that our polynomial-time approximate method is a $2$-approximation to our exponential-time exact algorithm.
    \item To further ease the use of the new method, we present extensions of the algorithms, that can automatically learn the \emph{right} values for the two input parameters, namely motif size $k$ and length $l$, to discover interesting motifs. This considerably reduces the time and effort needed in exploratory analysis.
    \item We perform extensive quantitative and qualitative evaluation of our new methods on six real-world and 25 semi-synthetic TS and compare them to four state-of-the-art competitors. We show that our approximate algorithm finds larger motif sets given the same distance threshold and motif sets with smaller pairwise distances given the motif set size $k$. We furthermore illustrate that $k$-Motiflets lead to motifs that are clearer and easier to interpret. Experiments show that our approximate algorithm is faster than any of the competitors.
\end{enumerate}

\section{Background and Definitions}~\label{sec:definitions}
\begin{figure*}[t]
    \centering
	\includegraphics[width=1.6\columnwidth]{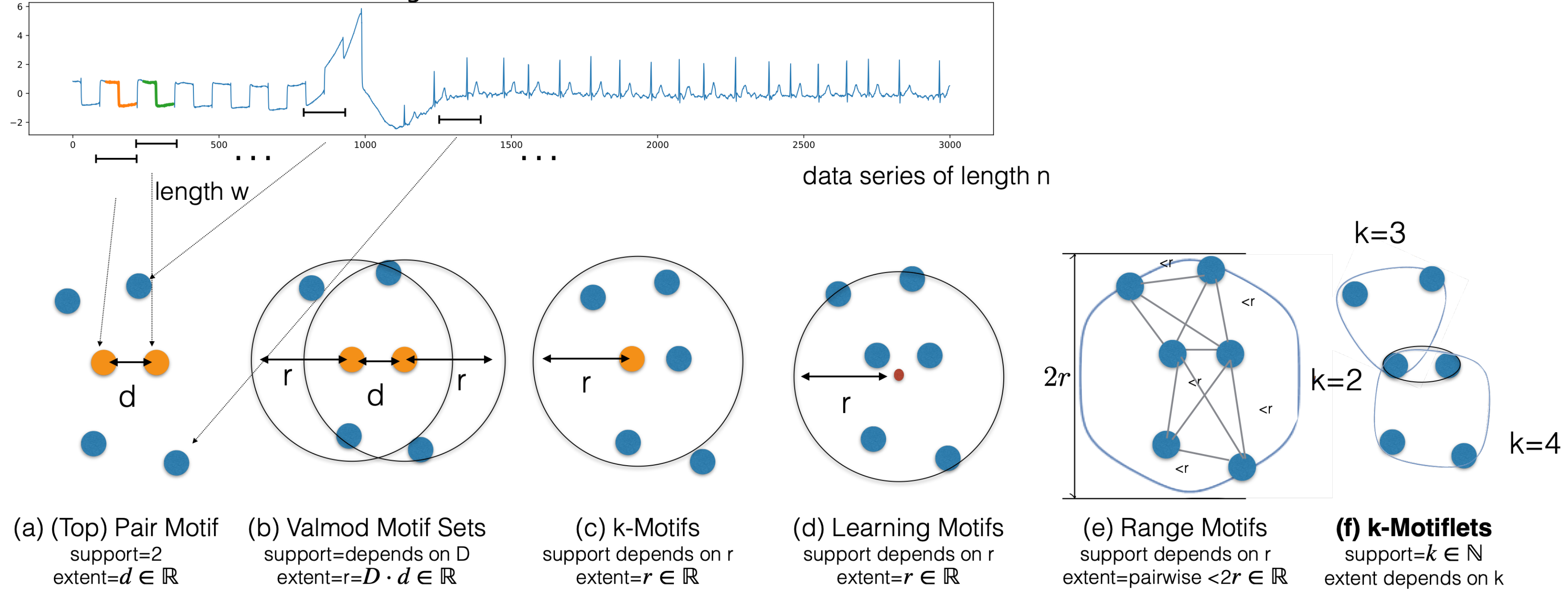}
	\caption{
	Illustration of the six different concepts of motif discovery. We consider six subsequences of fixed length as potential motif set and plot them in a 2-dimensional space. From left to right:
	(a) PM,
	(b) VS,
	(c) $K$-Motif,
	(d) LM,
	(e) RM,
    (f) $k$-Motiflets, for $k \in [2,3,4]$.
    Geometrically, VS, LM and $K$-Motif are (unions of) hyperspheres; RM and $k$-Motiflets are Reuleaux polygons.
	\label{fig:latent_motifs}
	}
\end{figure*}
In this section, we first formally define time series (TS) and the z-normalized Euclidean distance, which we (like all prior work) will use throughout this work. Next, we introduce the four existing MD definitions and show their differences using a geometric metaphor. We then introduce $k$-Motiflets by derivation from Range Motifs (RM), and relate them to the prior work.

\begin{definition}
\emph{Time Series}: A time series $T = (t_1, t_2,\dots, t_n)$ of length $n$ is an ordered sequence of $n$ real-values $t_i \in \mathbb{R}$.
\end{definition}


\begin{definition}
\emph{Subsequence}: A subsequence $S_{i;l}$ of a time series $T = (t_1, \dots, t_n)$, with $1 \leq i \leq n$ and $1 \leq i+l \leq n$, is a time series of length $l$ consisting of the $l$ contiguous real-values from $T$ starting at offset i: $S_{i;l} =(t_i,t_{i+1},...,t_{i+l-1})$.
\end{definition}

Works in MD typically exclude overlapping subsequences from consideration for motifs, as their distance is naturally low.

\begin{definition}
\emph{Overlapping subsequences (Trivial Match)}: Two subsequences $S_{i;l}$ and $S_{j;l}$ of length $l$ of the same time series $T$ overlap iff they share at least $l/2$ common offsets of $T$: $ (i - l/2) \leq j \leq (i + l/2) $.
\end{definition}

In the context of MD, the similarity of two subsequences is (almost exclusively) measured using the \emph{z-normalized Euclidean distance}.

\begin{definition}
\emph{z-normalized Euclidean distance (z-ED)}:
Given two subsequences $S_l^{(1)}=(s^{(1)}_1, \ldots, s^{(1)}_l)$ with mean $\mu^{(1)}$ and standard-deviation $\sigma^{(1)}$ and $S_l^{(2)}=(s^{(2)}_1, \ldots, s^{(2)}_l)$ with $\mu^{(2)}$ and $\sigma^{(2)}$, both of length $l$, their z-normalized Euclidean distance (z-ED) defined as:
\begin{equation}\label{eq:z-ED}
\textit{z-ED}(S_l^{(1)}, S_l^{(2)})=\sqrt{\sum_{t=1}^{l}\left(\frac{s^{(1)}_t - \mu^{(1)}}{\sigma^{(1)}} - \frac{s^{(2)}_t - \mu^{(2)}}{\sigma^{(2)}}\right)^2}
\end{equation}
\end{definition}

Using this distance function, we may now introduce a notion for the approximate matching of subsequences as basis for MD.

\begin{definition}
\emph{r-match}: Two subsequences $S_{i;l}$ and $S_{j;l}$ of $T$ are called $r$-matching iff (a) $z\text{-}ED(S_{i;l}, S_{j;l}) \leq r \in \mathbb{R}$ and (b) they are \emph{non-overlapping}.
A set $S$ of subsequences of $T$ is called $r$-matching, iff all subsequences in $S$ are pairwise $r$-matching.
\end{definition}

\subsection{Pair Motifs and Motif Sets}~\label{sec:pair_and_set_motifs}
We next define the two basic approaches to MD: \emph{pair motifs} and \emph{motif sets}. A geometrical and intuitive explanation of their differences is shown in Figure~\ref{fig:latent_motifs}. Note that the original definitions of LM, VS and K-Motifs left it undefined whether the subsequences in a motif set must be pairwise non-overlapping or not. As such, the $K$-Motif reference implementation~\cite{lonardi2002finding} discovers sets of pairwise overlapping subsequences. In this work, we require (by definition of r-matching) all sequences in a motif set to be non-overlapping.

\begin{definition}
\emph{Top Pair Motif (PM)}~\cite{mueen2009exact}: The top pair motif of length $l \in \mathbb{N}$ of $T$ is the pair of non-overlapping subsequences of length $l$ of $T$ with minimal distance.
\end{definition}

Obviously, two PMs may share the same distance. To solve ties when enumerating PMs, the PMs are typically returned in the order of appearance in the TS. Next, we present the four different definitions of motif sets that capture approximately repeated subsequences of a TS.

\begin{definition}\label{def:KM}
\emph{Top K-Motif}~\cite{patel2002mining}: Given radius $r \in \mathbb{R}$ and length $l \in \mathbb{N}$, the top K-Motif is the largest set $S$ of subsequences of length $l$ of $T$ for which the following holds: There exists a subsequence $S_{i,l}$ of $T$ which is $r$-matching to all members of $S$. We call $S_{i,l}$ the core of $S$, and $S$ the motif set (or just motif).
\end{definition}

Note that this definition requires that the core of the motif itself is a subsequence of $T$. This constraint is lifted in the following definitions; in the latest and most liberal definition, i.e., the range motif (see below), actually no core must exist anymore.

\begin{definition}\label{def:LM}
\emph{Top Learning Motif (LM)}~\cite{grabocka2016latent}: Given a time series $T$, radius $r \in \mathbb{R}$ and length $l \in \mathbb{N}$, the top LM of $T$ is the largest set $S$ of subsequences of length $l$ of $T$ for which holds: There exists a core sequence $C$ which is $r$-matching to all members of $S$.
\end{definition}

The only difference between k-Motifs and Learning Motifs is that for the latter the core of $S$ must not be a subsequence of $T$ itself. For this reason, LM motifs are also called latent motifs~\cite{grabocka2016latent}.

An alternative definition for MD was introduced by VALMOD.

\begin{definition}\label{def:VS}
\emph{VALMOD Motif Set (VS)}~\cite{linardi2018valmod}: Given a time series $T$, its pair motif $S_{i;l}, S_{j;l}$, distance $r\in \mathbb{R}$ and length $l \in \mathbb{N}$, the VS of $T$ is the set of subsequences that are $r-matching$ to $S_{i;l}$ or $S_{j;l}$.
\end{definition}

Thus, the VS always computes the top pair motif first and then iteratively extends its two subsequences with close neighbours. Accordingly, the pairwise distances of subsequences in a VS motif may be up to $3r$.

The to-date most liberal definition of MD is that of~\cite{mueen2009exact}, which does not require a core to exist. While it is more than a decade old, no algorithm for computing such motifs has been published yet.

\begin{definition}\label{def:RM}
\emph{Top Range Motif (RM) Set}~\cite{mueen2009exact}: Given a time series $T$, distance $r\in \mathbb{R}$ and length $l \in \mathbb{N}$, the top RM of $T$ is the largest set of subsequences $S$ from $T$ that are pairwise $2r$-matching.
\end{definition}

Figure~\ref{fig:latent_motifs} shows a geometrical explanation for the differences between the five MD definitions based on the TS from Figure 1. For illustration,  we represent subsequences of $T$ as points in 2-dimensional space. Geometrically, the LM set forms a hypersphere with radius $r$ around a latent core $S_l$. K-Motifs form a hyper-sphere of radius $r$ around a core $S_{i;j}$ of $T$. VS is the union of two hyperspheres of radius $r$ around the pair motif. Finally, RM forms a so-called Reuleaux Polygon: A shape created by the union of circles of radius $r$ around the subsequences of the motif.

Observe that the circle of diameter $2r$ as defined by a LM is a special case of a Reuleaux Polygon and consequently of a RM. All points inside the LM have pairwise distance smaller than or equal to its diameter $2r$. In fact, RM returns the same shape as LM iff a circular shape covers the most subsequences among all possible shapes defined by Reuleaux Polygons. The opposite is not true, as a Reuleaux Triangle of width $2r$ cannot be covered by a circle with diameter $2r$. Therefore, the RM is a more general definition than LM; nevertheless, it is still under-researched. The computational complexities of finding the exact LM or RM are unknown. Solving PM and K-Motifs problem is quadratic in the TS length $n$. A list of existing implementations can be found in Table~\ref{table:discovery_algs}.

\subsection{k-Motiflets}
All of the aforementioned MD definitions have in common that their motif sets depend on two parameters, i.e., the length $l$ of subsequences and the distance threshold $r$. Especially $r$ is very hard to set in practice, as it is very difficult to get an intuition regarding a threshold on the z-normalized distance of subsequences of a TS. Furthermore, already slight variations of its value may lead to grossly different motifs which makes tuning rather brittle. Yet, no methods for learning the parameters from the data are known. In contrast, $k$-Motiflets, which will be defined in the following, is independent of $r$. Instead, it requires users to set an integer parameter $k$ that defines the size of the motif set to be discovered. This measure is easy to understand, has much less possible values (integer versus real), can be learned from the data (Section~\ref{sec:extensions}), and inherently leads to smoothly growing motif sets in experimentation (see Figure~\ref{fig:elbow_method} for an example). Before introducing $k$-Motiflets, we first have to define the \emph{extent} of a motif.
\begin{definition}\label{def:extent}
\emph{Extent}: Consider a TS $T$ and a set $S$ of subsequences of $T$ of length $l$. The \emph{extent} of $S$ is the maximal pairwise distance of elements from $S$:
$$ d = extent(S) = \max_{(S^{(1)}, S^{(2)}) \in S \times S} (\textit{z-ED}(S^{(1)}, S^{(2)}))$$
\end{definition}
We next define $k$-Motiflets. These could actually build on any of the existing motif set definitions; we use RM to achieve maximal flexibility.
\begin{definition}\label{def:motiflet}
\emph{Top $k$-Motiflet}: Given a time series $T$, cardinality $k \in \mathbb{N}$ and length $l$, the top $k$-Motiflet of $T$ is the set $S$ with $|S|=k$ subsequences of $T$ of length $l$ for which the following holds: All elements of $S$ are pairwise $d$-matching, with $d=extent(S)$, and there exists no set $S'$ with $extent(S') < extent(S)$ also fulfilling these constraints.
\end{definition}
Note that the top $k$-Motiflet is not unique if two (or more) sets of k-subsequences share the same smallest distance. k-Motiflets have a blind spot for same $k$-retrieval. Given two motif sets of the same size, only the one with smaller extent is reported. Thus, in some pathological case, two motif sets of the same size $k$ may hide each other, if one motif always has a smaller extent than the other for all $k' \in [2 \dots k]$. In this case, our algorithm will only return the motif set with the smallest extent for each $k'$, creating the blind spot. In contrast, SotA has a blind spot for same $r$-retrieval. Given two motif sets of the same radius, only the larger one (in terms of $k$) is reported. Thereby, SotA hides distinct motifs given the same radius as input.
Geometrically, a $k$-Motiflet is the smallest Reuleaux polygon that covers $k$ subsequences. For the special case of $k=2$, $2$-Motiflets return the pair of subsequences with smallest distance, which is equal to the Pair Motif (PM) definition. For any $k\geq 2$ this represents the RM of size $k$ with smallest $r$. Figure~\ref{fig:latent_motifs} illustrates $2,3,4$-Motiflets in comparison to the other definitions.

\section{Related Work}\label{sec:related_work}
\begin{table*}
\small
\centering
\begin{tabular*}{2\columnwidth}{@{\extracolsep{\fill}}|c|c|c|c|c|}
\hline
\textbf{Motif Type} & \textbf{Name} & \textbf{Worst Case Complexity} & \textbf{Properties} & \textbf{Exact?} \tabularnewline
\hline
\hline
\multirow{3}{*}{Motif Pairs~\cite{mueen2009exact}} & MK~\cite{mueen2009exact} & $\mathcal{O}(ln^{2})$ & Admissible Pruning & Yes \tabularnewline
\cline{2-5}
\cline{2-5}
 & SCRIMP~\cite{zhu2018matrix} & $\mathcal{O}(n^{2})$ & Runtime independent of $l$ & Yes \tabularnewline
\cline{2-5}
 & VALMOD~\cite{linardi2018matrix} & $\mathcal{O}((l_{max}-l_{min})\cdot n^{2})$ & Variable length over ranges & Yes \tabularnewline
\hline
\multirow{5}{*}{K-Motifs~\cite{lonardi2002finding}} & EMMA~\cite{lonardi2002finding} & $\mathcal{O}(ln^{2})$ & SAX-based, produces trivial-matches & Heuristic\tabularnewline
\cline{2-5}
  & GrammarViz~\cite{senin2014grammarviz, senin2018grammarviz} & $\mathcal{O}(n^{2})$ & Discretization (SAX), variable length & Heuristic \tabularnewline
\cline{2-5}
  & ScanMK~\cite{bagnall2014finding} & $\mathcal{O}(ln^{2})$ & & Heuristic \tabularnewline
\cline{2-5}
  & ClusterMK~\cite{bagnall2014finding} & $\mathcal{O}(ln^{2})$ & Hierarchical Clustering & Heuristic\tabularnewline
\cline{2-5}
  & SetFinder~\cite{bagnall2014finding} & $\mathcal{O}(ln^{2})$ & & Heuristic \tabularnewline
\hline
Learning Motifs~\cite{grabocka2016latent} & Learning Motifs~\cite{grabocka2016latent} & $\mathcal{O}(ln)$ & Non-convex Gradient Desc. & Heuristic\tabularnewline
\hline
VALMOD Motif Sets~\cite{linardi2018matrix} & VALMOD~\cite{linardi2018matrix} & $\mathcal{O}((l_{max}-l_{min})\cdot n^{2})$ & Variable length & Exact \tabularnewline
\hline
Range Motifs~\cite{mueen2009exact} &  None & - & No known implementation. & - \tabularnewline
\hline
k-Motiflets &  k-Motiflets & $\mathcal{O}(k \cdot n^2 + n \cdot k^2)$ & Learn parameters $l$ and $k$ & Both \tabularnewline
\hline
\end{tabular*}
\caption{Overview of state-of-the-art Pair Motif and Motif Set discovery definitions and implementations, given a motif length $l$ and TS of length $n$. Notably, no Range Motif discovery algorithm was published to-date. There are four different formal MD definitions for Motif Sets we are aware of.\label{table:discovery_algs}}
\end{table*}

MD in TS has been researched intensively for approximately $20$ years. The first publication we are aware of was studied in the context of summarizing and visualizing massive TS datasets~\cite{lonardi2002finding}. In the following, we shall first discuss recent approaches to pair MD and then focus on methods for the discovery of motif sets.

The MK algorithm~\cite{mueen2009exact} from Mueen et al. published in 2009 is likely the most widely used baseline for pair MD. However, it is outperformed by more recent methods in terms of runtime, in particular QUICK MOTIF~\cite{li2015quick}, STOMP~\cite{zhu2016matrix}, SCRIMP~\cite{zhu2018matrix}, and VALMOD~\cite{linardi2018matrix}.
QUICK MOTIF first builds a summarized representation of the data using Piecewise Aggregate Approximation (PAA) and arranges these summaries in Minimum Bounding Rectangles within a Hilbert R-Tree index for pruning. STOMP and SCRIMP are based on the computation of the matrix profile~\cite{yeh2016matrix}, which represents the 1-nearest-neighbor (1-NN) subsequence to each subsequence of a TS. The subsequence pair with smallest distance among all is the motif pair. VALMOD~\cite{linardi2018matrix} addresses the limitation that previous works always assumed a user-defined motif length $l$. Instead, they proposed an efficient algorithm for finding best pairs within a range $[l_{min} \dots l_{max}]$. In our evaluation in Section~\ref{sec:experiments}, we shall use VALMOD for fixed length $l$ only. VALMOD was extended to return motif sets by performing a range-query around the two pair motif sequences. A common characteristic of algorithms for pair MD is a complexity of up to $\mathcal{O}(n^2 l)$, for a TS of length $n$ and a fixed motif length $l$ (compare Table~\ref{table:discovery_algs}). Using SCRIMP to compute the pairwise distance matrix, this may be reduced to $\mathcal{O}(n^2)$. Our implementation of $k$-Motiflets is based on the fast formulation of this problem as in SCRIMP~\cite{dokmanic2015euclidean,zhu2018matrix}, but extended for $k$-NN distances; it will be described in more detail in Section~\ref{sec:motiflets}.

EMMA~\cite{lonardi2002finding} was the first K-Motif discovery algorithm. It is based on the discretization of subsequences using SAX. In short, Symbolic Aggregate approXimation (SAX)~\cite{lin2007experiencing} transforms an input TS into a string (word) based on computing mean values over intervals, and the discretization of these mean values. The SAX words are then hashed into buckets, where similar subsequences hash into similar buckets, and the buckets are subsequently post-processed to obtain the final motif sets. Also GrammarViz~\cite{senin2014grammarviz,senin2018grammarviz} is based on SAX, on which it applies a linear-time algorithm Sequitur~\cite{nevill1997linear} for grammar inference. From the detected  rules, those are derived that represent reoccurring subsequences. Like EMMA, the method is heuristic, as both mine motifs in the discretized SAX space, which can lead to two similar subsequences being considered as different.

ScanMK, ClusterMK, and SetFinder have been proposed by the same authors~\cite{bagnall2014finding} as solutions to K-Motif discovery. ScanMK initializes set motif candidates with pair-motifs that are within a distance lower than $r$. From these two subsequences all nearby subsequences within $r$ are queried and added to the set motif candidate. Finally, the set is condensed to remove subsequences that are more than $2r$ apart. ClusterMK is based on a bottom-up hierarchical clustering of the best-matching pairs of clusters within distance $r$. First, the closest pairs of subsequences are merged to form initial clusters. A cluster is then represented by averaging its members. Clustering terminates once the distance between clusters is larger than $r$. SetFinder directly searches the $r$-matches of every subsequence and outputs the highest cardinally set.

The concept of Learning Motif (LM)  was introduced by Grabocka et al.~\cite{grabocka2016latent} to better deal with noisy TS. The paper approaches LM discovery as a process which, starting from a random initialization, iteratively modifies a motif core $S'$ to increase its frequency, i.e., the size of the surrounding motif, while keeping its radius fixed. As the frequency function is not differentiable, they propose a smooth Gaussian-kernel approximation that allows to use gradient ascent to find the hopefully best hidden motif cores. The LM solution is a heuristic, as the optimization problem is non-convex and the gradient ascent might get stuck in a local optimum.

Range Motif (RM) discovery was defined in~\cite{mueen2009exact}. It is the most liberal definition (see Section~\ref{sec:pair_and_set_motifs}), as it does not require a motif core to exist anymore. To-date, no algorithm has been published implementing this RM concept. However, $k$-Motiflets are based on RM, as for any $k \geq 2$ we return the RM with smallest $r$ to cover exactly $k$ subsequences (see also previous Section). Our algorithms for computing $k$-Motiflets could thus easily be turned into a solution for the original RM problem by running them for increasing values of $k$ until the distance threshold is violated.

\section{Exact and Approximate k-Motiflets}\label{sec:motiflets}
In this section we will present two algorithms solving the $k$-Motiflet problem. The first algorithm, presented in Section~\ref{sec:approximate_motiflets}, is an efficient heuristic with polynomial runtime. Furthermore, we show that it is a $2$-approximation of the exact solution. The second algorithm, presented in Section~\ref{sec:exact_algorithm}, is exact but has exponential runtime in $k$, and uses the heuristic solution as initial solution for pruning. Before describing the concrete algorithms, we first give an intuition of their inner working in Section~\ref{sec:intuition}.

Both algorithms expect parameters $l$ and $k$ to be given. While we share the necessity to set $l$ manually with all other methods except VALMOD, we replace the usual parameter $r$ (distance threshold) with $k$ (size of the motif set). Although $k$ is much easier to understand, setting it nevertheless might require time-consuming exploratory analysis. To reduce these efforts, we shall present two bespoke methods for learning both parameters $l$ and $k$ from the data in  Section~\ref{sec:extensions}.

\subsection{Intuition of Approximate Solution}\label{sec:intuition}
\begin{figure}[t]
    \centering
	\includegraphics[width=0.50\columnwidth]{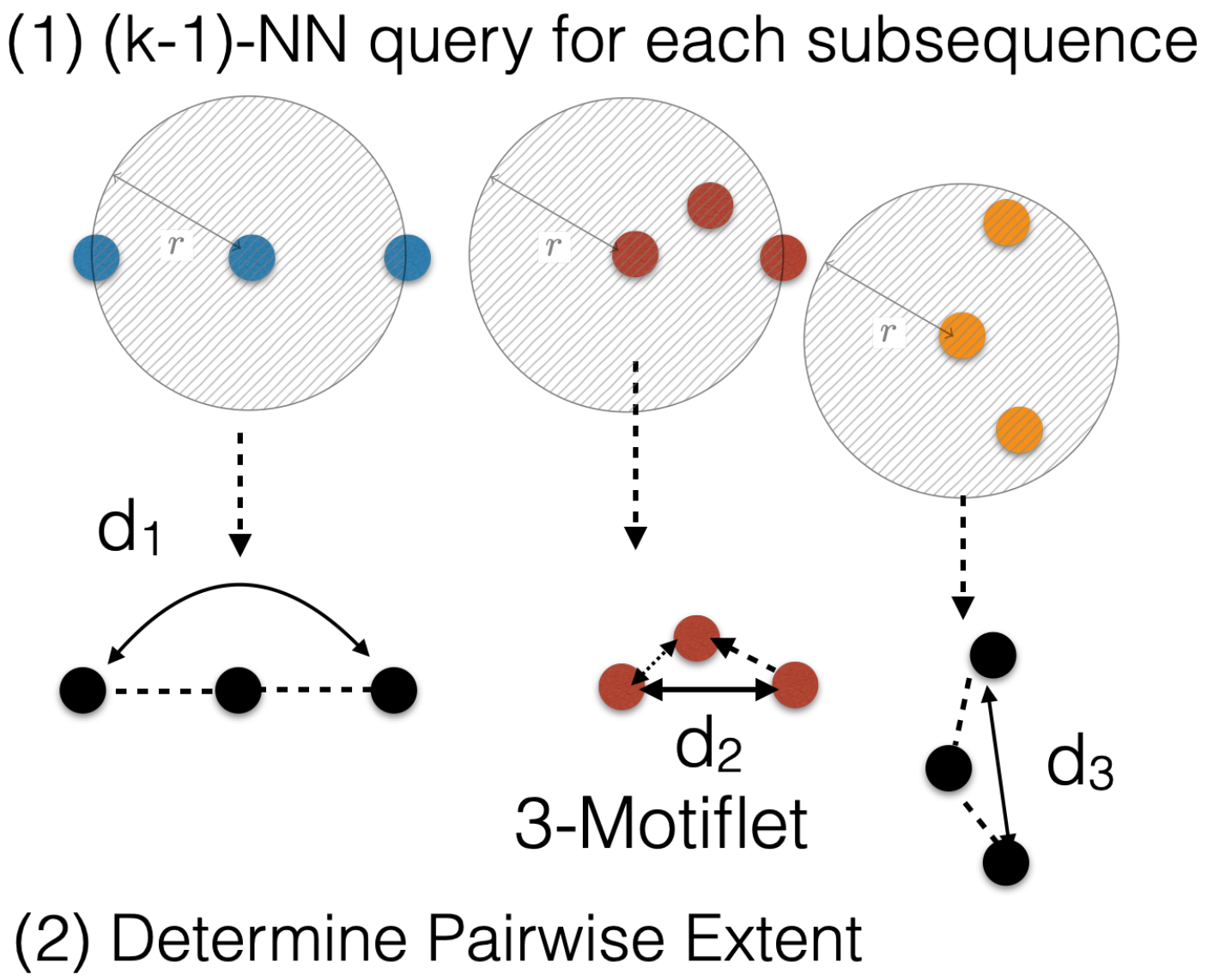}
	\caption{
	Depicted are three sets (blue, red, orange) with radius $r$ around a query. $k$-Motiflet discovery involves two steps: (1.) (k-1)-NN search around each query subsequence, and (2.) determine the extent of each set, i.e. $d_1, d_2, d_3$. Finally, the top $k$-Motiflet with smallest extent $d_2$ is returned.
	\label{fig:approximate_motiflet_discovery}
	}
\end{figure}

Given a motif length $l$ and a motif size $k$, the algorithmic idea of our algorithms for computing $k$-Motiflets is the following: we start by building motif set candidates by joining each subsequence of a TS $T$ with its non-overlapping $(k-1)$-NNs. Next, we compute for each of these sets of cardinality $k$ the \emph{extent}, i.e. the maximum over all pairwise distances. Note that the subsequences in each set must not be pairwise as similar to each other as both are to the core $S$, as two neighbours of $S$ may be on opposite sites of the hyperspace centered around $S$. Thus, to find the best k-Motiflet, we cannot simply pick the smallest $(k-1)$-NN distance, but must explicitly determine the motif set with smallest extent. This process is illustrated in Figure~\ref{fig:approximate_motiflet_discovery}, from $(n-1)$-NN search (top) to computing the extent (bottom) and choosing the motif set with smallest extent. Furthermore, without further modifications this method is a heuristic, as it only considers Motiflets built from the $k-1$-NNs of a core from $T$.

We will first outline an approximate solution to the $k$-Motiflet problem. We will show in Section~\ref{sec:quality} that this solution is a $2$-approximation and present an exact algorithm in~\ref{sec:exact_algorithm}.

\subsection{Approximate k-Motiflet Algorithm}\label{sec:approximate_motiflets}
\begin{algorithm}[t]
    \small
	\caption{Compute Approximate k-Motiflets}\label{alg:approx-k-motiflets}
	\begin{algorithmic}[1]
		\Procedure{get\_k\_motiflets}{$T$, $k$, $l$}
		\State $D_{i,j}$ $\gets$ \textsc{calc\_distance\_matrix}(T, l)
        \State $(motiflet, d) \gets (\{\}, inf)$ \Comment{bsf of extent}
        \For{$i \in [1, \dots, (\text{len(T)} - l + 1)] $}
            \State $idx \gets \textsc{argwhere}(D_{i} < d) \cup \{i\}$
            \If{$len(idx) \geq k$}
                \State $candidate$ $\gets$ \textsc{non\_trivial\_kNN}$(k, idx)$
                \State $dist$ $\gets$ \textsc{pairwise\_extent}$(D, candidate, d)$
                \If{$dist < d$}
                    \State ($d, motiflet) \gets (dist, candidate)$
                \EndIf
            \EndIf
		\EndFor

		\State \Return $(motiflet, d)$
		\EndProcedure
	\end{algorithmic}
\end{algorithm}

Algorithm~\ref{alg:approx-k-motiflets} takes as an input the TS $T$, the size of the motiflet $k \in \mathbb{N}$ and motif length $l \in \mathbb{N}$. First, we compute the pairwise z-normalized distance matrix (line~2). The algorithm applies admissible pruning to reduce the number of candidate sets using an upper bound on the best-so-far extent $d$ (see line~3). It iteratively checks if there are at least $k$ subsequences within a $d$-range (line~6). If \emph{true}, we extract the closest non-trivial $k$ subsequences (line~8), and determine the pairwise extent of this set (line~9). In $\textsc{PAIRWISE\_EXTENT}(\_, \_, d)$ we apply admissible pruning, too, by stopping the computation once any pairwise distance exceeds $d$. If the overall extent $dist$ is smaller than the best-so-far, we update the $k$-Motiflet (lines~10--13). Finally the $k$-Motiflet and its extent $d$ are returned.

The presented algorithm is greedy and approximates the extent of the optimal set of subsequences in line~8, assuming that the NNs of a query are also the ones in the $k$-Motiflet. In  $\textsc{NON\_TRIVIAL\_kNN}()$ we order the subsequences by their distance to the query and return the closest non-trivial neighbours.
Figure~\ref{fig:approximate_motiflet_discovery} illustrates the idea of the algorithm and the steps involved in computing a $3$-Motiflet. We iteratively perform two steps for each subsequence $q$: (1) search for the $(k-1)$-NN of $q$, and (2) determine their pairwise extent of the candidate set. Finally, the set with minimal extent is returned (in red).


\subsubsection{Complexity:}
The runtime of \textsc{get\_k\_motiflets}{($T$, $k$, $l$)} is dominated by the computation of the pairwise z-normalized distance matrix (line~2). Our implementation is based on an efficient formulation of this problem from~\cite{dokmanic2015euclidean,zhu2018matrix}, extended for $k$-NN distances. This requires only $\mathcal{O}(n^2)$-time, which is independent of the motif length $l$. Next, the algorithm iterates through all cells of the distance matrix in lines~5--6 with $\mathcal{O}(n^2)$-time. Checking for non-trivial matches in line~8 requires $k$-times searching for the minimum over one row of the matrix with a complexity of $\mathcal{O}(k \cdot n^2)$ over all $n$ rows. We can compute the maximum of $k$ pairwise distances in $\mathcal{O}(k^2)$ (line~9). Accordingly, the for-loop is in $\mathcal{O}(n \cdot k^2)$. Thus, the overall \emph{worst case} runtime complexity is:
$\mathcal{O}(k \cdot n^2) + \mathcal{O}(n \cdot k^2)$.
In the \emph{best case}, due to admissible pruning, the first subsequence (first row of the matrix) is the top $k$-Motiflet and we can prune all further computations in the first cell of each subsequent row. The \emph{best case} runtime complexity is thus:
$\mathcal{O}(n^2) + \mathcal{O}(k^2)$.
\subsubsection{2-Approximation}~\label{sec:quality}
Algorithm~\ref{alg:approx-k-motiflets} only computes an approximate solutions, as it only considers Motiflets built from the $(k-1)$-NNs of a core from $T$, whereas the top $k$-Motiflet may not contain this core. In the following, we will first show that our method precisely is a $2$-approximation, for $k=3$, by constructing a worst-case instance, and then extend it to the case of $k>3$.

\begin{figure}[t]
    \centering
	\includegraphics[width=0.8\columnwidth]{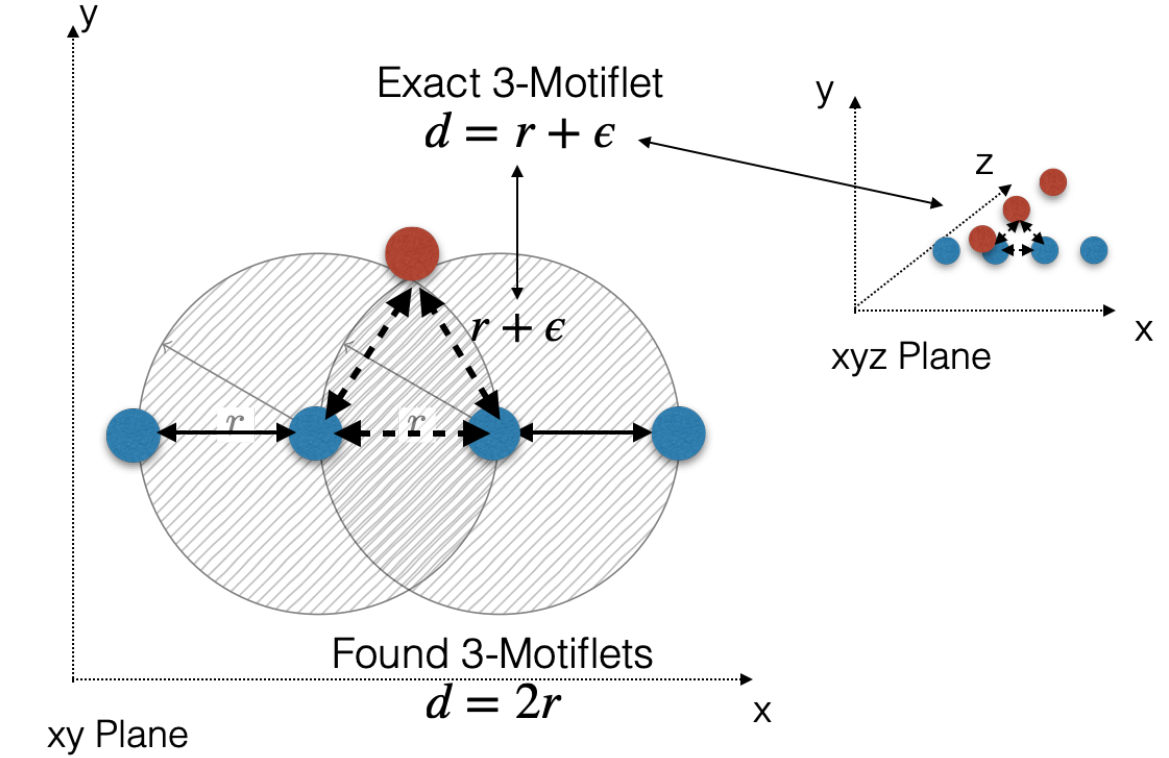}
	\caption{
	Worst case example for k=$3$ on xy and xyz-plane. Our approximate algorithm will return a $3$-Motiflet with extent $2r$. However, the optimal $3$-Motiflets has extent $r+\epsilon$.
	\label{fig:adverse_examples}
	}
\end{figure}

\emph{Case $k=3$}: Figure~\ref{fig:adverse_examples} illustrates such a worst case example for 3-Motiflets in both the xy-plane (for ease of illustration) and xyz-plane. The blue dots and red dots represent subsequences that are equally distributed on a grid with an offset of $r$. The offset between the red dot and the blue dots in the xy-plane shall be $r+\epsilon$, for an arbitrarily small $\epsilon \in \mathbb{R}^+$.
In the case of $3$-Motiflets our approximate algorithm searches for $2$-NNs of each subsequence, which in this example are always one unit of $r$ away (illustrated by the hyperspheres). Thus, the pairwise extent is at most $d=2r$ for two subsequences on the diameter. However, the optimal $3$-Motiflet can be seen in the center of the figure: for each of the blue dots, the first $2$-NNs are $r$ away, and only their $3$-NN (red dot) is $r+\epsilon$ away. Thus, the top $3$-Motiflet has a pairwise distance of $d=r+\epsilon$, and consists of two blue dots and one red dot.
This example is also the worst case instance for $k=3$. The factor is largest when $\epsilon$ is close to 0, maximizing the difference of $2d$ and $\epsilon +r$. Increasing or decreasing epsilon will reduce the factor of the approximation as it is either not a part in the $k$-Motiflet $\epsilon>r0$ or is found by our approximate algorithm (for $\epsilon<0$). The worst case instance is thus the case, when the red dot touches the intersections of the circles.
\emph{Case $k>3$}: We now extend this worst case example to $k>3$ (using $L_1$ distance for the sake of simplicity): Assume we have a subsequence in $n \geq k+2$-dimensional space $S=(0, 0, \dots, 0)$ with $k-1$-neighbours $B^{i}, i\in[1,\dots,k-1]$, where only the i-th dimension is set to $r$. E.g.: $B^{1}=(r,0,\dots, 0)$ and $B^{2}=(0,r,\dots, 0)$.
$S$ and all $B^{i}$ constitute a $k$-Motiflet with extent $d=2r$.
Now, we hide a single subsequence $R_0=(r/2+\epsilon, r/2+\epsilon, 0, \dots, 0)$ at the intersections of the hypersphere (as in Figure~\ref{fig:adverse_examples} for $k=3$). $R_0$ has a distance of $r+\epsilon$ to $S$ and $B_1$. Finally, we add $k-1$ additional subsequences $R^{i}, i\in[3,\dots,k+1]$, where the first two dimensions are set to $r/2+\epsilon$ and the i-th dimension is set to $r$. E.g. $R^{3}=(r+\epsilon, r/2+\epsilon, r, \dots, 0)$. These are all distance $r$ away from $R_0$, and at least $2r$ away from $S$.

Thus, the top $k$-Motiflets consists of the subsequences $R_i$ and $S$ with total extent $r+\epsilon$, but the found $k$-Motiflet has extent $2r$ with $B_i$ and $S$.
Accordingly, our algorithm is a $2$-approximation of the exact solution. Note that $2$ is the maximal error; we actually observe much smaller factors on our real-world use cases, typically lower than $1.1$ also for much higher $k$ (see Section~\ref{sec:comparison_exact_approx}).


\subsection{Exact k-Motiflets}~\label{sec:exact_algorithm}
The approximate algorithm gives an upper bound on the extent $d$ of the optimal $k$-Motiflet solution. The algorithmic idea of our exact algorithm is based on an enumeration of all subsets of subsequences of $T$ of size $k$ combined with aggressive pruning. The pruning is based on the observation that the $k$-Motiflet must be within range $d$ of a subsequence $S$ in TS, as all other subsequences are within pairwise distance of at most $d$ to $S$. To obtain the overall smallest set in terms of its extent, we hence may prune the $d$-range candidate sets to return the smallest $k$-element set. However, enumerating all subsets of $k$ subsequences from a set of $\hat{k}>k$ elements is a combinatorial problem, which has exponential growth in $\mathbb{O}(\hat{k}^k)$.

Algorithm~\ref{alg:exact_motiflets} applies admissible pruning by using the best-so-far extent $d$. It initializes $d$ by the result of our approximate algorithm in line~3. We iterate over all subsets of subsequences and return those within $d$-range. For choosing candidate subsequences, we have to check each subset of size $k$ and compute its extent (line~7--9). Overall, there are $\mathbb{O}(n^k)$ in the worst case. The $k$-element motif set with lowest extent is the top $k$-Motiflet.

Thus, if we compare the approximate (Algorithm~\ref{alg:approx-k-motiflets}) and the exact solution (Algorithm~\ref{alg:exact_motiflets}), the main difference is in the enumeration of the candidate subsequences. Unfortunately, the exact algorithm has an exponential growth in the worst case, which makes it infeasible for even small $k$'s, which we will show in the experimental section. There are also good reasons to believe that the $k$-Motiflet problem is NP-hard, but a formal proof still has to be found. However, we will show that our approximate $k$-Motiflets algorithm gives good results in our experimental evaluation in Section~\ref{sec:experiments}.

\begin{algorithm}[t]
    \small
	\caption{Compute Exact k-Motiflets}\label{alg:exact_motiflets}
	\begin{algorithmic}[1]
		\Procedure{get\_exact\_k\_motiflets}{$T$, $k$, $l$}
		\State $D_{i,j}$ $\gets$ \textsc{calc\_distance\_matrix}(T, l)
        \State $(motiflet, d) \gets \textsc{get\_k\_motiflets}(T, k, l)$ \Comment{approximation}
        \For{$i \in [1, \dots, (\text{len(T)} - l + 1)] $}
            \State $idx \gets \textsc{argwhere}(D_{i} < d)  \cup \{i\}$
            \If{$len(idx) \geq k$}
                \For{$k\_subset \in $ \textsc{non\_trivial\_subsets}$(k, idx)$}
                    \State $dist \gets$ \textsc{pairwise\_extent}$(D, k\_subset, d)$
                    \If{$dist < d$}
                        \State ($d, motiflet) \gets (dist, k\_subset)$
                    \EndIf
                \EndFor
            \EndIf
		\EndFor

		\State \Return $(motiflet, d)$
		\EndProcedure
	\end{algorithmic}
\end{algorithm}




\section{Parameter Selection}~\label{sec:extensions}
In this section, we present methods to automatically find suitable values for the motif length $l$ and set size $k$ so, that meaningful concealed structures of an input TS are found without domain knowledge. No comparable method exists for any of the competitor definitions. As MD in TS is an unsupervised problem, we cannot claim to find optimal values under all circumstances. For instance, the methods we present will not produce meaningful results when applied to entirely random TS, as in those simply no "suitable" $k$ or $l$ exist at all. However, we found them to be very effective and helpful in our experiments (Section~\ref{sec:experiments}).
Our methods for determining values for $k$ and $l$ are based on an analysis of the extent function:
\begin{definition}
\emph{Extent Function (EF)}: Assume a fixed length $l$ and a time series T. Let $S_k$ be the top $k$-Motiflet with length $l$ of $T$. Then, the \emph{extent function} $EF$ for $T$ is defined as $EF(k) = extent(S_k)$.
\end{definition}

\begin{figure}[t]
    \centering
	\includegraphics[width=1.0\columnwidth]{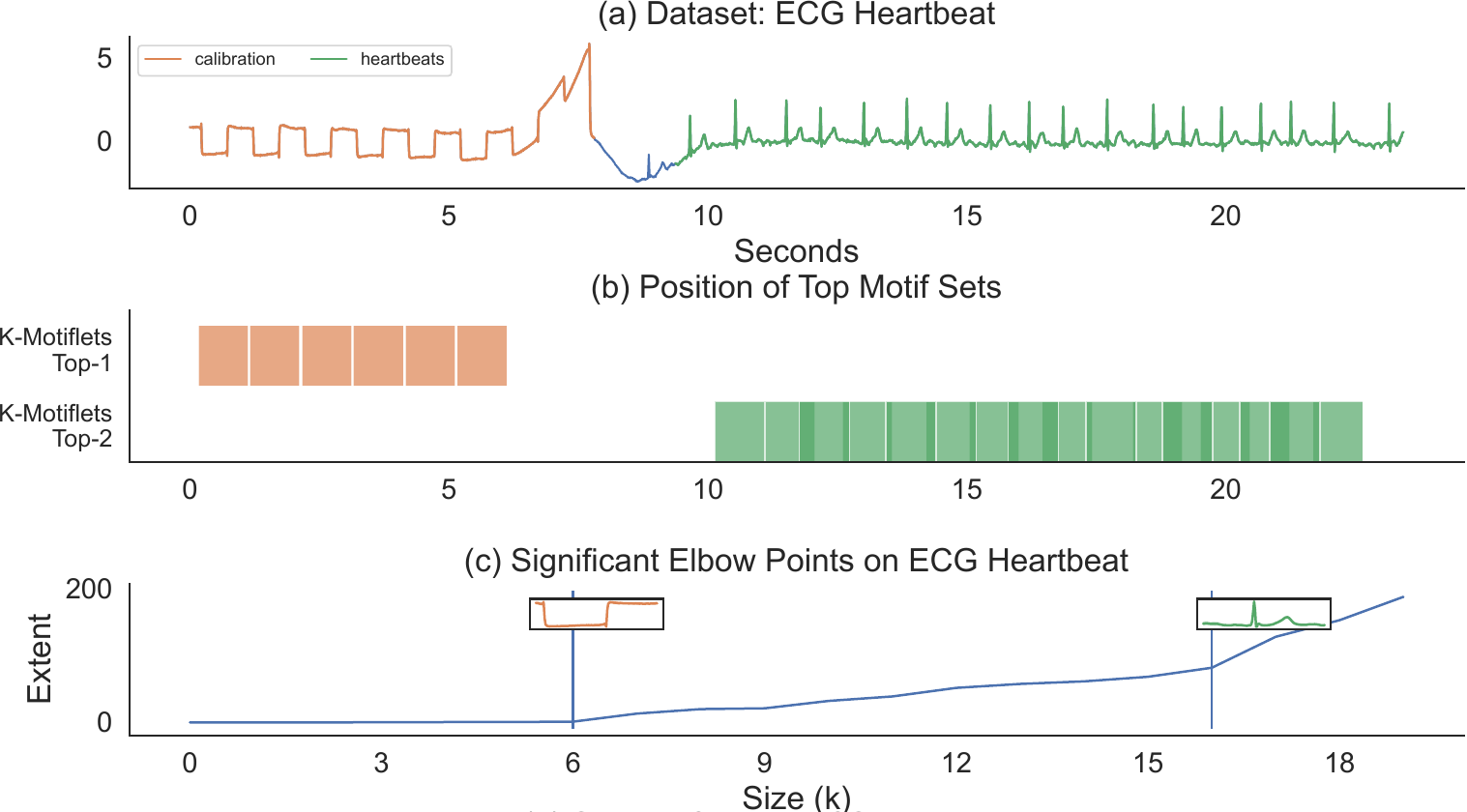}
	\caption{
	The \emph{EF} (bottom) is a function of the cardinality of $k$-Motiflets to the extent. Elbow points represent large changes in similarity of the found motif, indicative of a concept change from calibrations signal to heartbeats. The runtime was $0.5$s to compute the EF.
	\label{fig:elbow_method}
	}
\end{figure}

Note that EF can be efficiently computed when starting from the largest value to be considered, as $extent(S_{k+1})$ is a (usually rather tight) bound for $extent(S_k)$, which allows for aggressive pruning in all cases but the first. When considering the curve of the EF, we can observe the following: First, EF is monotonically increasing in $k$. Second, if the slope of $EF$ increases \emph{slowly} from $k$ to $k+1$, then $S_k$ can be extended to $S_{k+1}$ with a motif that is very similar to the subsequences of $S_k$. A longer interval $[k,\dots,k+n]$ for which EF increases only slowly and that cannot be extended (i.e., $EF(k-1)$ is considerably smaller than $EF(k)$ and $EF(k+n+1)$ is considerably larger than $EF(k+n)$) very probably stem from a set of $n+1$ subsequences of $T$ whose $k$ most similar elements build $S_k$ and where for every increment of $k$ another highly similar subsequence from $T$ exists and is added to build the next top motiflet.
On the other hand, if $EF(k+1)-EF(k)$ is large, i.e., if we have a steep increase between $k$ and $k+1$ - we call  this an \emph{elbow point} - then very probably the $k$-Motiflet could not be exceeded with further, highly similar subsequences. In such cases, the $k+1$ motiflet very likely is formed by an entirely new motif - with more occurrences, but at the price of a larger extent. Examples for both observations, i.e., long flat stretches and elbow points, can be found in Figure~\ref{fig:elbow_method}. These considerations lead to the following two ideas:
(a) Elbow points in the \emph{EF} indicate changes of motifs. The last value of $k$ before the elbow indicates a maximal motiflet, which we consider a particularly meaningful value for $k$. We study these points in Section~\ref{sec:k_selection}.
(b) Long flat stretches of the EFs indicate a high number of occurrences of a motif, but depend on the motif length $l$. Accordingly, we consider values of $l$ leading to long flat stretches as particularly meaningful. Section~\ref{sec:window_size_selection} describes how we find such values.

\subsection{Learning meaningful k}~\label{sec:k_selection}
Elbow points, i.e. points with a notable increase in slope of the \emph{EF} between two values $k$ and $k+1$, indicate that the $k$-Motiflet cannot be extended to $k+1$ without a strong increase in its extent. Consider Figure~\ref{fig:elbow_method} as an example. We set the motif length to $1$ sec, i.e. 60 bpm. We observed the following: the \emph{EF} (Figure~\ref{fig:elbow_method} bottom) is flat until 6 repetitions of the calibration signal have been found. This stretch ends with a notable increase in slope until $k=16$, where another elbow point exists. These two points are characteristic for this TS. The corresponding motifs are depicted at the bottom of the figure, with $6$ and $16$ occurrences, respectively.

We tested different methods for finding elbow points, such as~\cite{satopaa2011finding} or scipy. We found that a simple threshold $\alpha$ on the slope of the \emph{EF} performed best for the particular shapes that $EF$ typically exhibit. Given a list of $k-1$ extents $p=[d_i]_{i=2}^k$, we thus define elbow points as follows:
\begin{alignat*}{2}
    elbow(p(i), p(i+1)) &=
        \left\{\begin{array}{lr}
            1, & \textit{iff } m1 / m2 > \alpha \\
            0, & \textit{else} \\
        \end{array}\right\} \\
    \textit{with }
        m1 &= p(i+1)-p(i)+\epsilon \textit{, } \\
        m2 &= p(i)-p(i-1)+\epsilon
\end{alignat*}
A small constant $\epsilon \in \mathbb{R}^+$ is added to the slope to avoid dividing by $0$. An elbow is found, if the slope rises by a factor of $\alpha$.

\emph{Perquisites and Limitations:}
Reliably detecting elbows is a hard challenge. Our choice of $\alpha$ relies on the assumption that the data contains motif sets. These create visible elbows on all use cases (compare Figure~\ref{fig:quantative_analysis_1}) . Yet, if this assumption is violated, such as in random walk data, every elbow detection method is  cursed to fail. We would expect such random series to contain no reasonable motif sets, and, in fact, the elbow function is a rather straight. Still, our method reports “elbows” at the points where the line is not completely straight. Some more research is needed to investigate this. However, in our experiments the change in extent was always so significant, that we choose to default this value to $\alpha=5$, and all experiments in this paper use this default. Yet, the user may choose different elbow points without extra computations. Further evaluations of learn-k on more data sets can be found on~\cite{MotifletWebPage}.

\subsection{Learning Motif Length l}~\label{sec:window_size_selection}
As described before, the length of a flat stretch in the \emph{EF} corresponds to a maximal motiflet. The existence and lengths of such stretches depend on $l$. For instance, if we set $l$ exactly to the periodicity of the heartbeat or of the calibration wave in Figure 2, the $EF$ contains two long flat stretches each corresponding to a (high) number of occurrences of the respective motif. Learning a suitable motif length thus can be approached by searching values for $l$ that create long flat stretches in the $EF$.

Accordingly, one could find $l$ by computing EFs for a range of $l$ values and chose the one where the EF contains the longest flat stretch. This, however, would have two drawbacks: (1) we would assume that only a single motif exists, and (2) stretches of different lengths may also exhibit different (small) slopes and are thus hard to compare, i.e., given a single threshold for "flat" would be difficult. Instead, we determine $l$ by calculating a normalized \emph{area under the EF}, abbreviated as $EF_AU$, as steeper stretches or smaller stretches - necessarily ending with an elbow point and thus an increase in slope - lead to larger areas under $EF$.

For a given length $l$, let $EF^{(l)}$ be the list of the $k-1$ extents $p^{(l)}=[d^{(l)}_i]_{i=2}^k$ and $e^{(l)}$ be the number of identified elbows. The \emph{$AU\_EF$} score for length $l$ is defined as:
\begin{alignat*}{2}
    AU\_EF(p^{(l)}) &= \frac{1}{e^{(l)}} \sum_{i = 2}^k \frac{(d^{(l)}_i - min(p^{(l)}))}{(max(p^{(l)})-min(p^{(l)}))}  \in [0, 1] \\
    \textit{ and  }
    best &= \min_{l \in [l\_min, l\_max]} AU\_EF(p^{(l)})
\end{alignat*}

\begin{figure}[t]
    \centering
    \includegraphics[width=0.5\columnwidth]{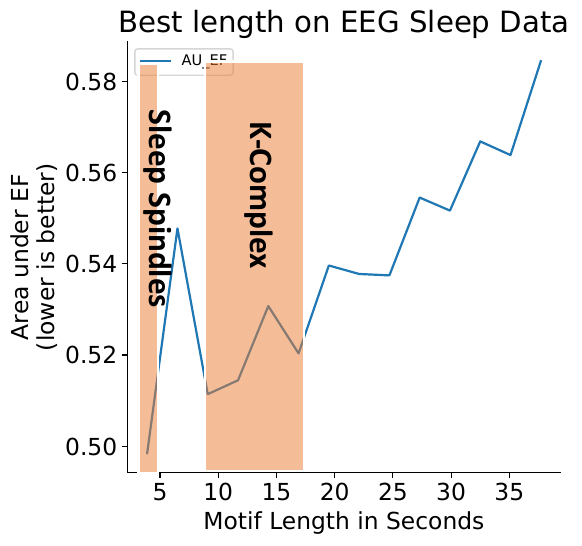}\includegraphics[width=0.5\columnwidth]{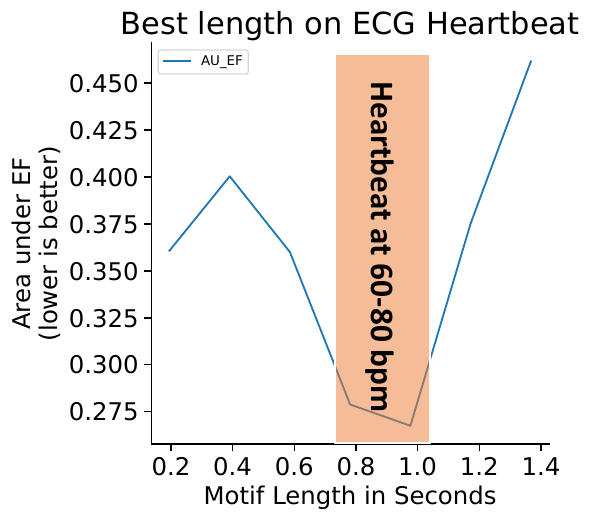}
	\caption{
	The \emph{Area Under the EF (AU\_EF)} captures the frequency of approximate repeats. The minima roughly capture known motifs in the two datasets, corresponding to the sleep spindles and K-Complex in sleep data (left) and heartbeats at a rate of $60$ to  $80$ bpm in ECG data (right).
	\label{fig:window_size_selection}
	}
\end{figure}
I.e., in our implementation we iterate over reasonable values for $l$ and choose the value where $AU\_EF()$ is minimal.

Figure~\ref{fig:window_size_selection} shows two plots of the  \emph{AU\_EF} for a range of discrete motif lengths $l \in [25, \dots, 200]$ for our running ECG example (left) and also an EEG sleep data set (right). The AU\_EF has its minimum around $l=0.8s$ to $l=1s$, equal to a heartbeat rate of $60$-$80$ bpm. The resulting motif sets are shown in Figure~\ref{fig:elbow_method}. In EEG sleep data the local minima correspond to sleep spindles at $5s$ and K-Complexes at $10-20s$ in sleeping cycles. Further evaluations of this method on more data sets can be found on our supporting web page~\cite{MotifletWebPage}. On the presented two datasets and ranges, learn-l took $1.5$s for ECG and $18.6$s on EEG. Runtimes vary with input ranges $[l\_{min}, l\_{max}]$.

\subsubsection*{Learning $k$ and learning $l$ for SotA}
As other methods expect parameter $r$ (not $k$) as input, we would need to implement a \emph{learn-r} routine, i.e. sample some $r$-values over an interval, compute the corresponding size $k$, and find elbow points. However, we observed that in our experiments the scores computed by the competitor algorithms were non-monotonically increasing with $r$, due to the heuristic nature of SotA. I.e. there typically is no elbow to find. Instead, the size of the motif set sometimes suddenly drops with increasing radius - compare experimental results in Figure~\ref{fig:quantative_analysis_1}. We thus think that finding optimal parameters for other methods than ours remains an interesting future research direction.

\section{Experimental Evaluation}\label{sec:experiments}
Our experimental evaluation is three-fold: Firstly in Section~\ref{sec:quantitative_analysis} we compare our \textbf{approximate} algorithm against SotA in a quantitative analysis on six real world sets. We evaluate methods by the similarity and cardinality of motif sets. Secondly in Section~\ref{sec:qualitative_analysis}, we compare our \textbf{approximate} $k$-Motiflet algorithm against SotA in a qualitative analysis on three real world and 25 semi-synthetic data sets - those for which motifs are known. We evaluate methods by their ability to find these motifs. Finally in Section~\ref{sec:comparison_exact_approx}, we compare our \textbf{approximate and exact} $k$-Motiflets regarding quality of results, and both against competitors for runtimes.

\emph{Competitors:} We compare our new algorithm to VALMOD~\cite{linardi2018valmod}, EMMA~\cite{lonardi2002finding}, Set Finder (SF)~\cite{bagnall2014finding}, and Learning Motifs (LM)~\cite{grabocka2016latent}. We used the original implementation of EMMA and SF but performed some runtime optimizations. In EMMA we improved the computation of means and stds from $\mathbb{O}(nl)$ to $\mathbb{O}(n)$ using a sliding window implementation. For SF, we improved the filtering of trivial matches by sorting the indices first and then searching for overlaps in sorted indices, rather than checking all pairs of indices. We used our own python re-implementation of VALMOD because the reference implementation is for pair motifs only. For LM, we used the JAVA implementation provided by the authors.

\emph{Data Sets:} We collected six challenging real-life and generated 25 semi-synthetic data sets. For four out of these, the literature describes the existence of motifs though without actually annotating them, see Table~\ref{tab:use-cases} for an overview.
\emph{Muscle Activation} was collected from in-line speed skating~\cite{morchen2007efficient} on motor driven treadmill with EMG data of movements. It contains  $29.899$ measurements at $100Hz$, equal to $30s$. Known motifs are muscle movement and a recovery phase.
\emph{ECG Heartbeats} contains a patient's (ID $71$) heartbeat from the LTAF database~\cite{petrutiu2007abrupt}. It contains $3.000$ measurements at $128Hz$, equal to $~23s$. The heartbeat rate is $60$ to $80$ bpm. Known motifs are a calibration signal and the heartbeats.
\emph{Vanilla Ice - Ice Ice Baby:} This dataset was used as introductory example. Its famous riff has $20$ repeats of roughly $3.6$ to $4$s and is $231$s long.
\emph{Physiodata - EEG sleep data} contains a recording of an afternoon nap of a 20 to 40 years old person~\cite{kohlmorgen2000identification}. Data was recorded with an extrathoracic strain belt. It consists of $269.286$ points at $100hz$, equal to $45min$. Known motifs are sleep spindles and $K$-Complexes.
\emph{Industrial Winding Process} records a plastic web being unwound from an unwinding reel, over a traction reel, and finally rewound on a rewinding reel~\cite{bastogne1997application}. Recordings correspond to the traction of the second reel's angular speed. It contains $2.500$ points sampled at $0.1s$, corresponding to $250s$.
\emph{Functional near-infrared spectroscopy (fNIRS)} contains brain imaginary data recorded at $690nm$ intensity. There are $208.028$ measurements. It contains four motion artifacts, due to movements of the patient, which dominate MD~\cite{dau2017matrix}. 
\emph{25 x Synthetic (Hexagon/UCR):} We took the first $25$ TS from the Hexagon challenge~\cite{dau2019ucr} and implanted, at random offsets, motif sets of sizes $k \in \{5,\dots,10\}$ and length $l=500$ using a \emph{bell shape}. We recorded ground truth motif size $k$ and radius $r$ for each.

\begin{table*}
\small
\centering
\begin{tabular}{|c|c|c|c|c|}
\hline
& Length & Known Motifs & Motif Length $l$ & Range for $k$\tabularnewline
\hline
\hline
Muscle Activation & 29899 (30s) & \makecell{Activation and Recovery} & Known: \textasciitilde ~$120$ ms & \makecell{ $ks=[2\dots20]$} \tabularnewline
\hline
ECG Heartbeats & 3000 (23s) & Calibration and Heartbeats & Known: \textasciitilde ~$0.8-1$ s & \makecell{$ks=[2\dots20]$}\tabularnewline
\hline
EEG Sleep Data & 269286 (45 min) & Sleep Spindles and K-Complex & Known: 1-6s and 30s & \makecell{$ks=[2\dots20]$}\tabularnewline
\hline
Ice Ice Baby & 23095 (231s) & Riff (20x) & Known: \textasciitilde ~$4$ s & \makecell{$ks=[2\dots20]$}\tabularnewline
\hline
\makecell{Industrial Winding} & 2499 (250s) & None & Learned: $l \in \{2,3,\dots,15\}$ & \makecell{$ks=[2\dots12]$} \tabularnewline
\hline
fNIRS & 208028 & None & Learned: $l \in \{1000,1200,\cdots,4000\}$ & \makecell{$ks=[2\dots20]$}\tabularnewline
\hline
25 x Synthetic (Hexagon/UCR) & 7500-15000 & Implanted Motifs & Known: $l = 500$, $k=[5\dots10]$ & \tabularnewline
\hline
\end{tabular}
\caption{Properties of real world use cases.\label{tab:use-cases}}
\end{table*}

\emph{Setting parameters $r$, $l$, and $k$:} A direct comparison of $k$-Motiflets to competitors is impossible, due to different parameterizations: Competitors require motif length $l$ and radius $r$, whereas $k$-Motiflets require length $l$ and the number $k$ of motif occurrences. Regarding $l$, we used the known value, if a value was known (see Table~\ref{tab:use-cases}), and otherwise learn $l$ as described in Section~\ref{sec:window_size_selection}. In any case, the value of $l$ was the same for all methods in all experiments. In contrast, the method for setting $k$ / $r$ are necessarily different each evaluation.

\emph{Hardware:} All scalability experiments ran on a server running LINUX with 2xIntel Xeon E5-2630v3 and 64GB RAM, using python version 3.8.3.
\emph{Reproducibility:} To ensure reproducible results we provide source codes and results on our website~\cite{MotifletWebPage}.
\subsection{Quantitative Analysis}\label{sec:quantitative_analysis}
\begin{figure}[t]
    \centering \includegraphics[width=1\columnwidth]{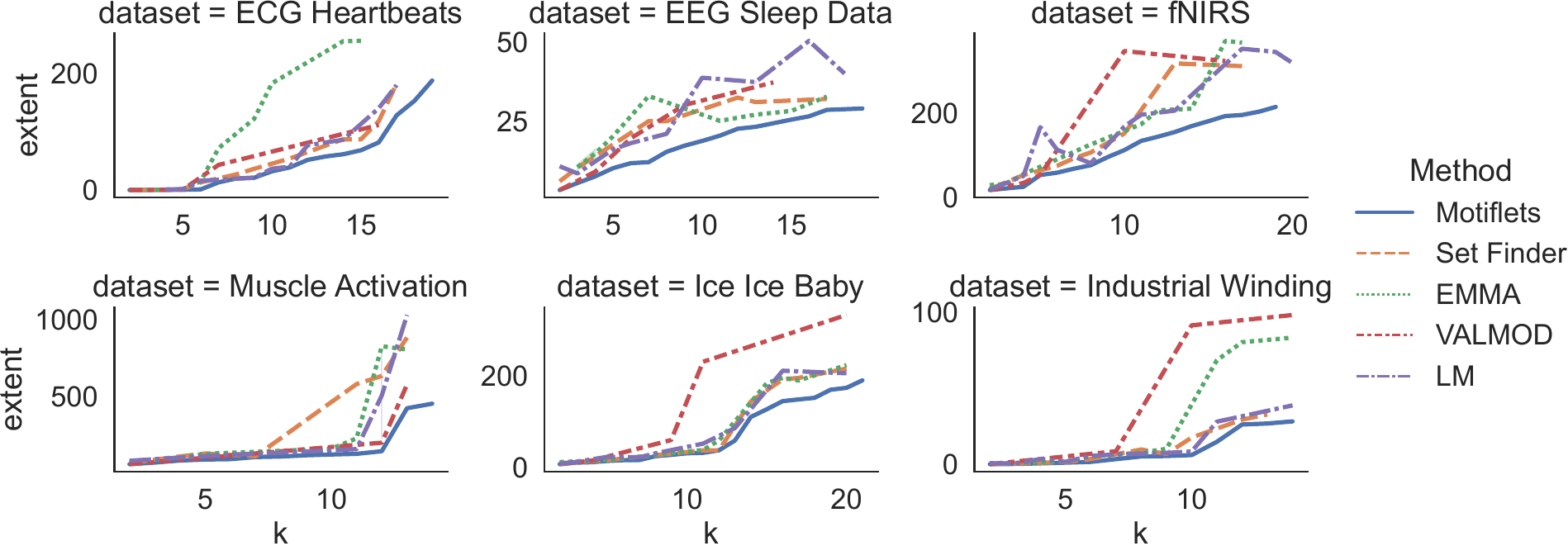}
	\caption{
    Relationship between cardinality and extent of the motifs. The curves of $k$-Motiflets are always below all competitors, e.g., it returns the largest motif set with highest similarity on all datasets by far.
	\label{fig:quantative_analysis_1}
	}
\end{figure}

We first compare the results of the approximate $k$-Motiflet algorithm to that of four state-of-the-art competitors using the six real TS. For these comparisons we performed an unbiased computation of extents and cardinalities of found motif sets at equivalent values of $r$ (respectively $d=2  \cdot r$) and $k$. In this evaluation, we find an MD method $M_1$ better than an MD method $M_2$, when $M_1$ finds larger motif sets at the same radius, smaller radii for the same motif cardinality, or both. The meaningfulness of the motif sets as found by different methods will be discussed in Section~\ref{sec:qualitative_analysis}.
We first ran each competitor for increasing values of $r$ and counted the cardinality $k$ and real extent of the found top motif sets, generating pairs of $(k,r)$. We then ran $k$-Motiflets for increasing values of $k$ and measured the extent of the found top motif set(s). which generates comparable pairs. Finally, we plotted the achieved extents by growing cardinality for each method. A good method finds motif sets with small extents even with increasing cardinalities, i.e. its line would be rather flat (parallel to x-axis). Figure~\ref{fig:quantative_analysis_1} shows that $k$-Motiflets in this regard shows the best performance of all methods. For each value of $k$ in each data set, it finds a motif set that has smaller than or at least equally small extent as all competitors; in turn, for each possible extent, it finds a larger or at least equally large motif set. Despite its distinct definitions, there is no clear second place among the other competitors, and we saw a different ranking on each of the use cases presented.
Recall that Valmod will return the Pair Motif for $k=2$. Thus, this experiment also shows that the special case of the $2$-Motiflets is equal to Pair Motif discovery, as the results are always equal to those of Valmod for $k=2$.

\subsection{Qualitative Analysis}\label{sec:qualitative_analysis}
In this section we discuss the quality of the discovered motif sets. The purpose is to compare methods not only by the size and extent of found motifs as in the previous section, but also to consider whether these motifs are actually meaningful, i.e., correspond to important events in the process producing the TS. We assume that a method finds a motif set if the reported motifs overlap with the ground truth.

\emph{Real-world datasets with Silver Standard Labels:} The literature mentions motifs for 4 of our 6 data sets. Yet, for none of them the precise motif occurrences are annotated, but only their rough shape and length. We used this information for creating a silver standard by exploiting $k$-Motiflets unique ability learn meaningful values for $k$ and $l$ (Section~\ref{sec:extensions}). Specifically, we learned values for $k$ and $l$ directly from the data, and compared the results with the descriptions from the papers. In all cases, the two Top-2 motifs corresponded very well to the descriptions. We compared the ability of the other methods to recover the respective motifs. We provided the competitors with proper values of $r$ and $l$, derived from the silver standard, but also with an added noise of $-10\%$ and $+10\%$ to reflect trial-and-error tuning. Note that in this setting our method has to recover meaningful motifs without any additional knowledge, while competitors are provided with close to optimal input values. Full results are shown on our webpage.

\emph{Ice Ice Baby by Vanilla Ice:} This song contains one famous motif set with $20$ repetitions roughly $4$s long from the introductory example. Learning-k (Section~\ref{sec:extensions}) took $3.4$s. Given these silver standard parameters, all competitor methods find this riff but with up twice as large extent. $k$-Motiflets is the only method to correctly find all $20$ repeats of the riff. All other methods include other subsequences into this motif, and with small noise added, this becomes worse.

\begin{figure}[t]
	\includegraphics[width=1.0\columnwidth]{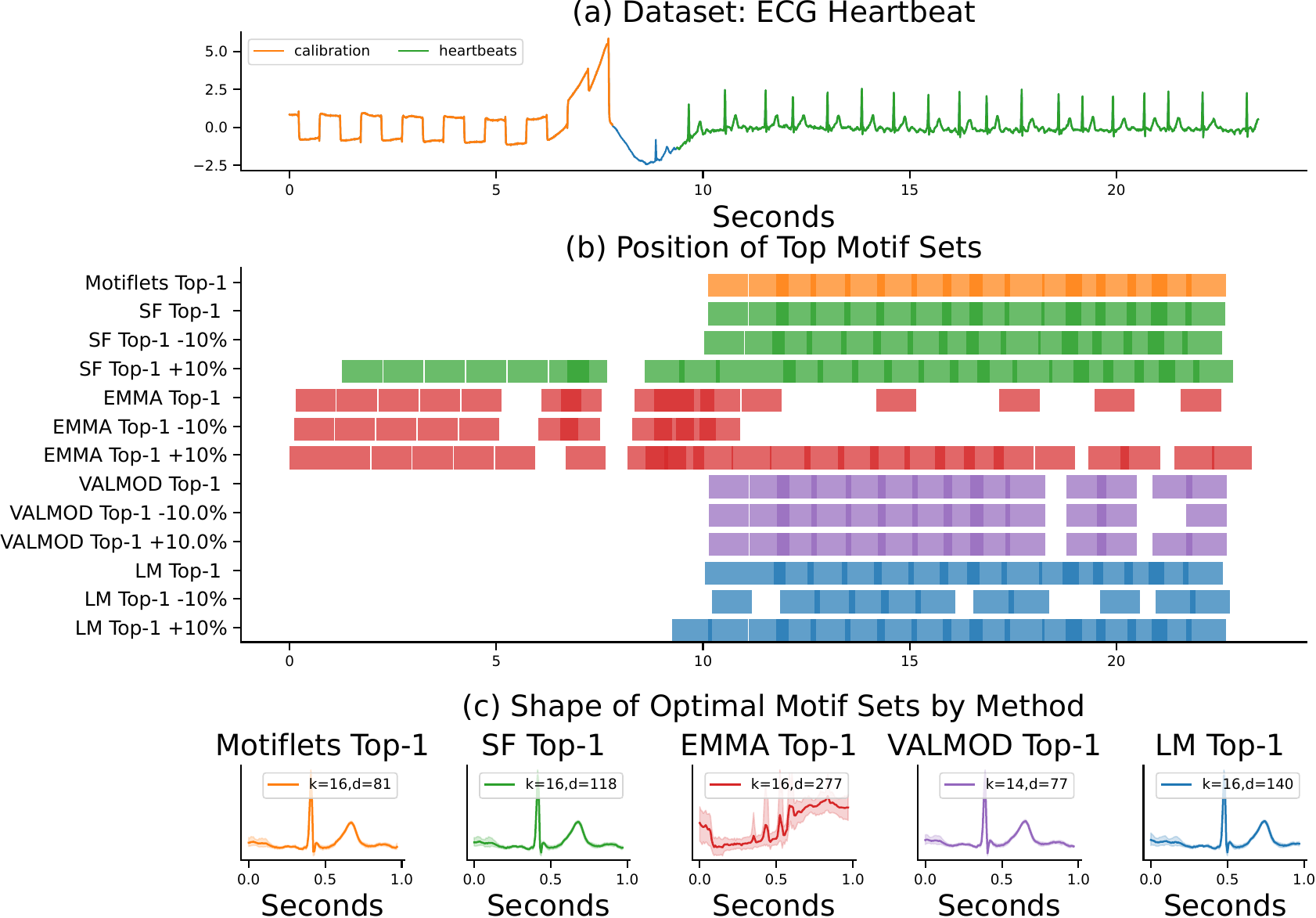}
 	\caption{The ECG trace~\cite{petrutiu2007abrupt} contains two motifs, starts with $6$  calibration signals, followed by $16$ heartbeats. For brevity, we show heartbeat results. Our approximate $k$-Motiflets identify all $k=16$ heartbeats. Only SF and LM find all occurrences of the heartbeats with optimal parameters, too, and adding noise results in blurred or too small motifs.
	\label{fig:set_motif_example}
	}
\end{figure}

\emph{ECG Heartbeats:} This data set was used throughout this paper. It contains two top motif sets, namely calibration and heartbeats (Figure~\ref{fig:set_motif_example}). We discuss only the top-1 motif, and our webpage shows the full results~\cite{MotifletWebPage}. Learn-l took $1.5$s, and learn-k took $0.5$s. Our k-Motiflet algorithm then identified $16$ heartbeats. Given silver standard parameters, only SF and LM find the top-1 motif, too, and accuracy depends heavily on the precise parameterization. EMMA only returns a blurred calibration signal.

\begin{figure}[t]
    \centering
    \includegraphics[width=1.0\columnwidth]{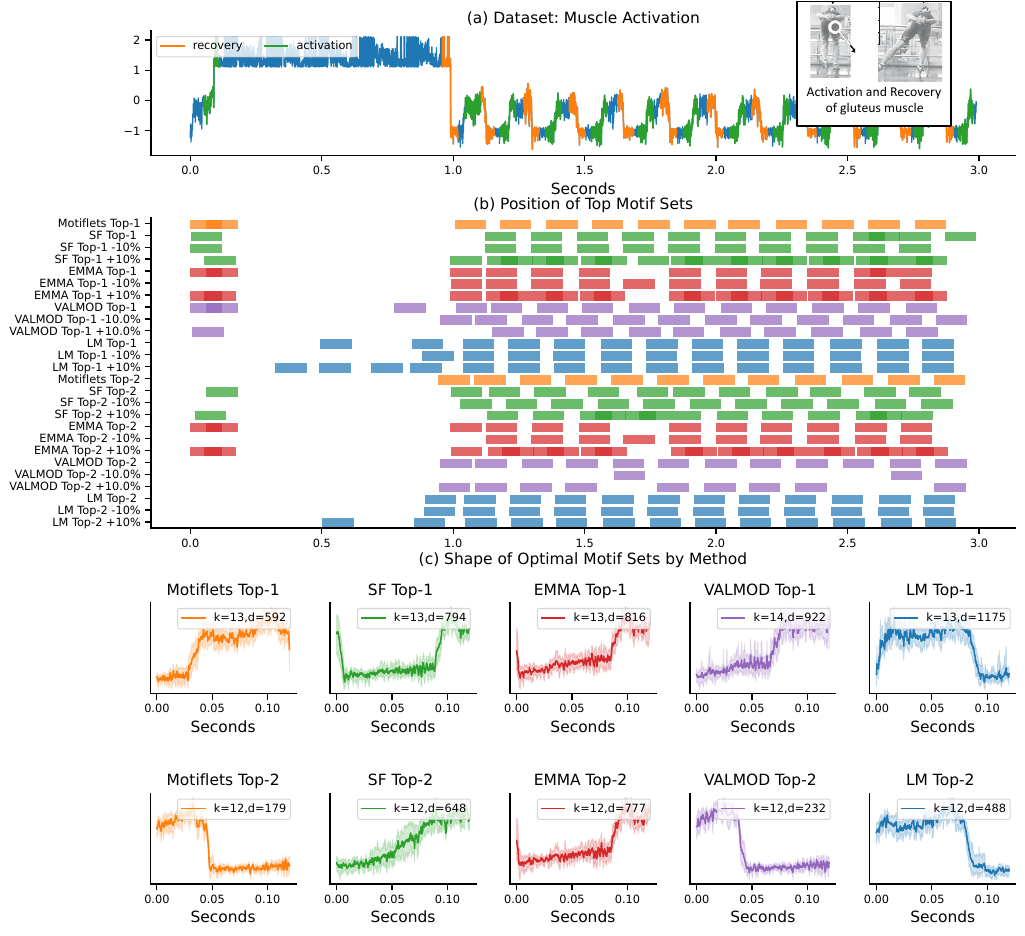}
    \caption{
    Top-2 motif sets for Muscle Activation~\cite{morchen2007efficient}.
    The top-1 motif set found by the approximate $k$-Motiflets alg. corresponds to the \emph{activation phase} and the top-2 motif to the \emph{recovery phase}. All methods find the \emph{activation phase}, but with up to $100\%$ larger extent. Valmod and LM found the recovery phase, again with up to $100\%$ larger extent.
	\label{fig:muscle_activation_top_1}
	}
\end{figure}

\emph{Muscle Activation:} The two top motifs present in this dataset are the activation (top-1) and the recovery phase (top-2) of the Gluteus Maximus muscle and have $13$ and $12$ occurrences, respectively. Learn-l took $5.0$s and learn-k took $3.9$s.
Given silver standard parameters, all competitor methods find the activation phase, e.g. the found motif set overlaps with the actual motifs, but with up to $100\%$ larger extent. $k$-Motiflets and  VALMOD are the only to identify both the activation phase as top-1 motif and recovery phase as top-2 motif but their extent $d=794$ is at least $75\%$ larger than that of $k$-Motiflets ($d=592$). The other methods identify the activation phase either as top-1 and top-2 motif. With small noise added, SF and EMMA stay locked on the activation phase, VALMOD alternates between both motifs.

\begin{figure}[t]
    \centering
	\includegraphics[width=0.5\columnwidth]{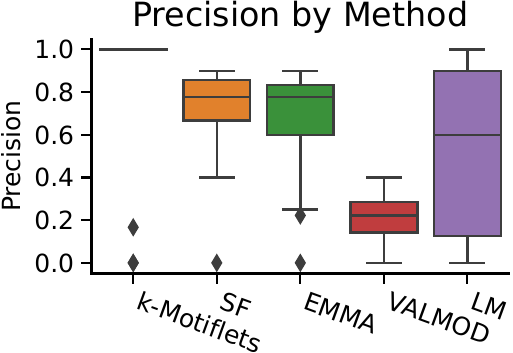}\includegraphics[width=0.5\columnwidth]{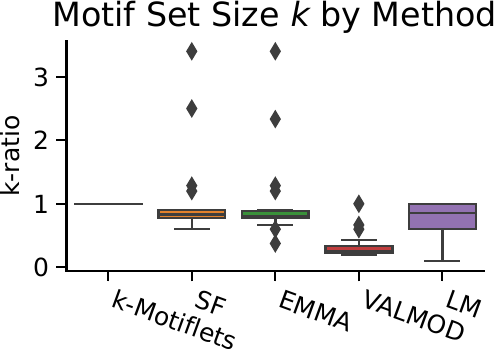}
	\caption{\emph{25 semi-synthetic datasets with ground truth}:
	Precision (left) and size (right) of the found motif sets by MD method. $k$-Motiflets performs best. 
	\label{fig:synthetic}}
\end{figure}

\emph{Semi-Synthetic Data Sets with Gold Standard Labels:} To measure the precision of the different MD methods we generated a semi-synthetic $25$ dataset benchmark from~\cite{dau2019ucr} with implanted motif sets. For each method, we used the gold standard parameters as inputs, i.e. the size $k$ for $k$-Motiflets or the radius of the implanted motif set $r$ for the competitors. Figure~\ref{fig:synthetic} shows the results as (a) precision, measured as overlap of reported methods with actual positions, and (b) the ratio of size of the result to the size of the implanted motif. Left: using only the size of the motif set, $k$-Motiflets have a precision of $100\%$ on all but $3$ datasets. Yet, even with exact range $r$ the competitors struggle to find the implanted motif set. The best competitors are SF and EMMA with far inferior median precision around $75\%$. Right: Note, that SF and EMMA report far larger motif sets for some datasets. The reason is the same as for $k$-Motiflets to fail on three datasets: there is a motif set present in the datasets of the same or larger size $k'$ as the implanted size $k$, but with a smaller overall radius $r'$ as the implanted radius $r$, and we have an arguably wrong gold standard. Thus, this present motif set hides the implanted motif set for $k$-Motiflets and leads to higher reports for SF and EMMA. Implanting motif sets with a guaranteed gold standard itself is a non-trivial problem, which we will address in future work. k-Motiflets outperform its competitors, as the hypersphere defined by SotA  (compare Figure~\ref{fig:motiflets-vs-sota} and Figure~\ref{fig:latent_motifs}) is always overestimating the actual hypersphere of the TOP motif set. E.g. the query radius around existing sequences is up to twice as large as needed by k-Motiflets, returning many false positives.
\subsection{Runtimes and exact k-Motiflets} ~\label{sec:comparison_exact_approx}
In the previous section, we evaluated the quality of the approximate $k$-Motiflet algorithm compared to four state-of-the-art MD methods. We did, however, not yet consider the exact $k$-Motiflet algorithm, because (a) its runtime is exponential in the size of the motif set and thus probably infeasible for larger values of $k$, and (b) we did not expect the motif sets found by the approximate version to be much worse than that found by the exact version. In this section, we experimentally verify both of these assumptions and subsequently also compare their runtimes to that of all competitors.

\subsubsection*{Scalability of approximate and exact $k$-Motiflet algorithms:}\label{sec:scalability}

\begin{figure}[t]
    \centering
    \includegraphics[width=0.5\columnwidth]{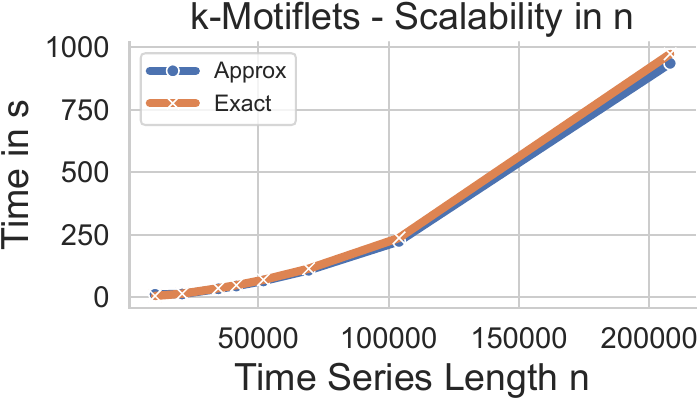}\includegraphics[width=0.5\columnwidth]{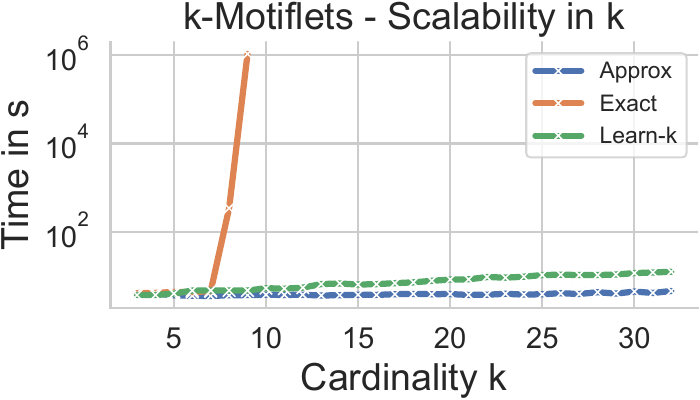}
	\caption{
	Runtime of the approximate vs. exact algorithms with motif length $l=100$. Left: as a function of the TS length $n$, with fixed cardinality $k=5$. Right: as a function of $k$ with fixed TS length $n=10000$. Both scale quadratic in $n$ but the exact algorithm is exponential in $k$.
	\label{fig:scalability_n}
	}
\end{figure}

\begin{figure}[t]
    \centering
	\includegraphics[width=1\columnwidth]{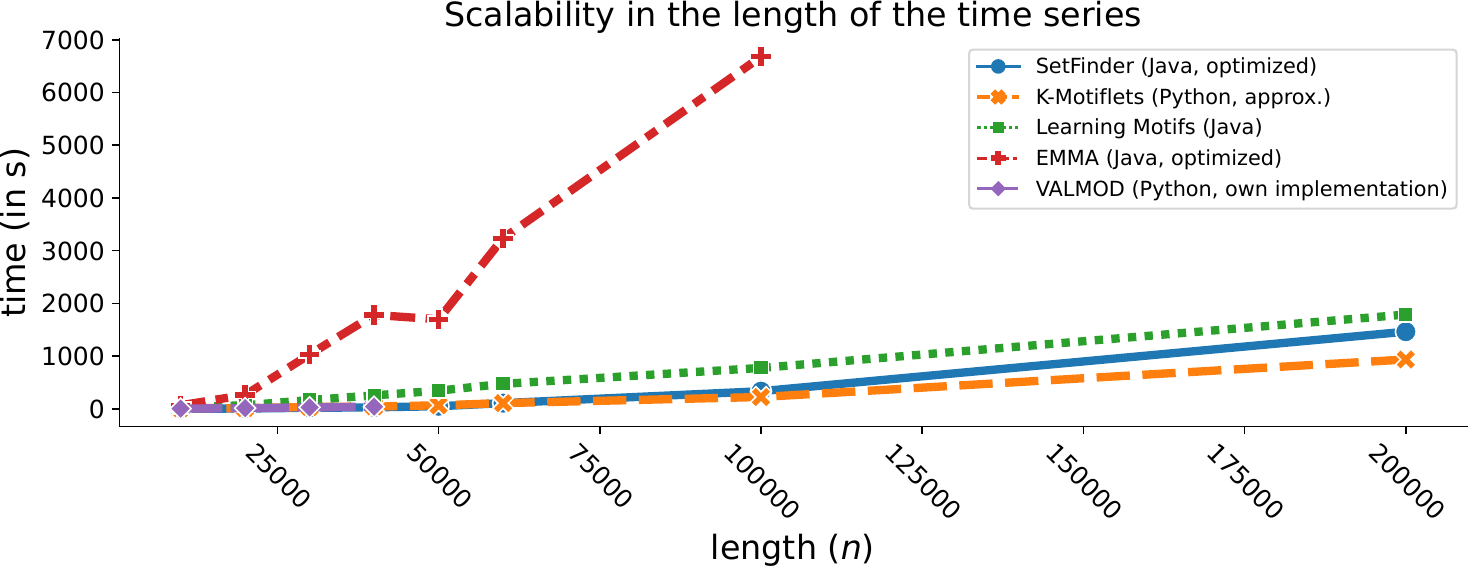}
	\caption{
	Scalability of different MD approaches in length n, with motif length $l=100$. $k$-Motiflets is fastest.
	\label{fig:scalability_n2}
	}
\end{figure}

We first study the scalability of the approximate and the exact $k$-Motiflet algorithm regarding the length $n$ of a TS and the cardinality $k$ of the motif sets. To this end, we use the largest TS from our data sets (fNIRS), encompassing $n=269.286$ time points.

Figure~\ref{fig:scalability_n}~(left) shows runtimes with growing TS length $n$, measured with fixed $k=5$. Interestingly, the runtimes of both methods are almost equal, resulting in $~11$ minutes for the full TS. Figure~\ref{fig:scalability_n}~(right) shows runtimes for growing values of $k$ at a fixed length $n=10.000$. Exact and approximate algorithms differ extremely for values of $k>7$, where the exponential complexity of the exact version results in a steep increase of the runtime. For instance, the runtime for $k=7$ is $4$ seconds, $5.7$ minutes for $k=8$, and already close to $12$ days for $k=9$. Thus, the exact algorithm becomes untraceable for larger $k$, even with the admissible pruning we implemented. In contrast, the runtime of the approximate algorithm remains below $2$ minutes even for $k=30$. We further plot the runtimes of the learn-k algorithm from Section~\ref{sec:k_selection}. The runtime overhead introduced by learn-k is around a factor of 2-3, i.e. the runtime is $4.56$s vs $12.5$s at motif sets of size $k=30$.
\begin{figure}[t]
    \centering
	\includegraphics[width=0.5\columnwidth]{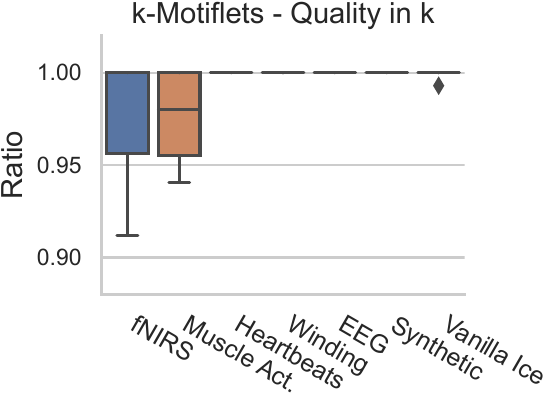}\includegraphics[width=0.5\columnwidth]{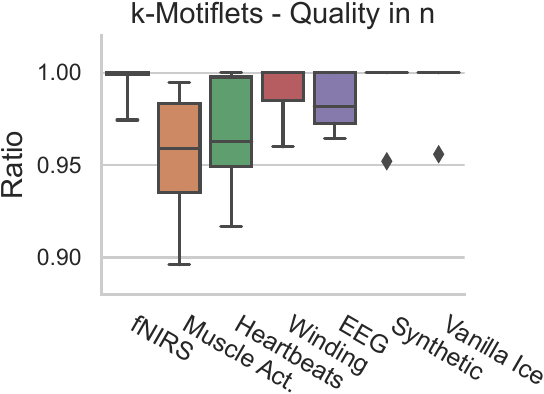}
	\caption{
	Quality as a ratio of the extents of the top-1 motif sets of the approximate to the exact algorithm. Left: Boxplot over ratio as a function of $k\in[2,...,9]$ with $n=10000$. Right: Boxplot over fractions of the full length n, $n' \in[1/8\%,1/7\%,\dots,100\%]$ with fixed cardinality $k=5$.
	\label{fig:scalability_k}
	}
\end{figure}
\subsubsection*{Quality of approximation:} We next studied the difference in results produced by the approximate and the exact $k$-Motiflet algorithms, depending on the length of a TS and on the chosen value of $k$. We first ran both algorithms on a growing prefix of all datasets for a fixed value $k=5$ and measured the ratio of the extents of the approximate to the exact solution over $n$ (i.e,. higher values mean that the approximate version is closer to the optimal solution). Results are shown in the box-plot in Figure~\ref{fig:scalability_k}~(right).
For some datasets when varying $k$, the approximate and exact solution always returned the same motifs, such as for heartbeats, window and EEG Figure~\ref{fig:scalability_k}~(left). Overall, we observed that the ratio is consistently over $91\%$ for varying $k$, and $89\%$ for varying $n$. This means that the approximate version finds motif sets close to the same extent as the exact algorithm.
We performed an in-depth analysis of the differences of results between the two methods. Interestingly, we found that they mostly come from cases where the parameter $k$ was set smaller than the actual number of  motif occurrences, which lead to the two methods finding different subsets of the same motif, which in turn led to slightly different extents. If the motif had e.g. $8$ occurrences, and we require $k=4$ occurrences, the approximate solution might find other $4$ (of the $8$) subsequences than the exact solution. When we move to the full frequency $k=8$, however, differences in extent faded away.

\subsubsection*{Runtime of competitors:} Finally, we compared the scalability of the approximate $k$-motiflet algorithm to its four competitors using the (largest) fNIRS dataset. For a fair comparison given different input parameters, we first set $k=5$, determined the extent of the top motif set as found by $k$-Motiflets, and used the corresponding radius as input for all competitors. We emphasize that the runtimes nevertheless are difficult to compare as the implementations use different languages (Java versus Python) and also show different efforts for runtime optimization. Under these circumstances, Figure~\ref{fig:scalability_n2} shows that our implementation of $k$-Motiflets is faster than all  competitor implementations we tested (despite being programmed in Python), though the differences to all methods except EMMA are rather small (less than factor 2).

\section{Conclusion}\label{sec:conclusion}
Often the first step in analyzing unlabelled TS is motif discovery, used to derive hypotheses from the data based on similar, frequent subsequences. However, existing tools for MD show a high variance in the discovered motifs depending on the given input parameter. If these parameters are set incorrectly this leads to the discovery of pure noise. In this paper, we presented a novel definition for motif set discovery, named $k$-Motiflets, which are the sets of subsequences with exactly $k$ approximate repeats and highest similarity in a given TS. We argued that the value of $k$ is much easier to set by a user than the usually used parameter $r$, which is the maximal similarity of a motif set. We presented an approximate and an exact algorithm for finding $k$-Motiflets and proved that the former is a $2$-approximation of the latter, which has exponential runtime in $k$. We also presented two algorithms along with our $k$-Motiflets for automatically learning appropriate values for $l$ and $k$ without any a-priori knowledge of the motifs.
By qualitative and quantitative evaluation on six real-world and 25 semi-synthetic use cases, we showed that the approximate algorithm produces better motifs than all its competitors at lower runtimes, and that its results come very close to the exact algorithm despite an exponentially lower runtime. Future work will consider variable length or multivariate motif discovery. We  envision new and better elbow detection methods in the case of time series where no motif set is present, violating our assumptions, and also addressing the blind spot of k-Motiflets detecting distinct motif sets of the same size.

\begin{acks}
We wish to thank Themis Palpanas, Rafael Moczalla, Arik Ermshaus, and Leonard Clauß for their input and fruitful discussions.
\end{acks}

\balance

\bibliographystyle{ACM-Reference-Format}
\bibliography{motiflets}

\end{document}